\documentclass[runningheads]{llncs}
\usepackage{graphicx}

\usepackage{tikz}
\usepackage{comment}
\usepackage{amsmath,amssymb} 
\usepackage{color}

\usepackage{graphicx}
\usepackage{amsmath}
\usepackage{amssymb}
\usepackage{booktabs}
\usepackage{fdsymbol}
\usepackage{ wasysym }
\usepackage{ stmaryrd }
\usepackage{units}

\usepackage{url}
\usepackage{hyperref}

\usepackage[utf8]{inputenc} 
\usepackage[T1]{fontenc}    
\usepackage{url}            
\usepackage{booktabs}       
\usepackage{amsfonts}       
\usepackage{nicefrac}       
\usepackage{microtype}      
\usepackage{xcolor}         

\usepackage{array}
\usepackage{bm}
\usepackage{dsfont}
\usepackage{multirow}
\usepackage{algorithm}
\usepackage{algpseudocode}
\usepackage[vlined,linesnumbered,ruled,algo2e]{algorithm2e}
\usepackage[linesnumbered,ruled,vlined]{algorithm2e}
\usepackage[flushleft]{threeparttable}

\usepackage{varwidth}
\usepackage{pifont}
\usepackage{makecell}
\usepackage{wrapfig}
\usepackage{enumitem}
\usepackage{caption}
\usepackage{dsfont}
\usepackage{soul}

\usepackage{listings}
\usepackage{color}

\begin{document}
\pagestyle{headings}
\mainmatter
\def\ECCVSubNumber{7441}  

\title{A Simple Single-Scale Vision Transformer for Object Detection and Instance Segmentation}

\titlerunning{Abbreviated paper title}
%

\author{Wuyang Chen\inst{1}\thanks{Work done during the first author's research internship with Google.} \and
Xianzhi Du\inst{2} \and
Fan Yang\inst{2} \\
Lucas Beyer\inst{2} \and
Xiaohua Zhai\inst{2} \and
Tsung-Yi Lin\inst{2} \and
Huizhong Chen\inst{2} \and
Jing Li\inst{2} \and
Xiaodan Song\inst{2} \and
Zhangyang Wang\inst{1} \and
Denny Zhou\inst{2}
}
\authorrunning{W. Chen et al.}
%
\institute{
University of Texas at Austin, Austin TX 78712, USA \\
\email{\{wuyang.chen,atlaswang\}@utexas.edu}\\
\and
Google\\
\email{\{xianzhi,fyangf,lbeyer,xzhai,tsungyi,\\huizhongc,jingli,xiaodansong,dennyzhou\}@google.com}
}

\maketitle

\begin{abstract}
This work presents a simple vision transformer design as a strong baseline for object localization and instance segmentation tasks.
Transformers recently demonstrate competitive performance in image classification. To adopt ViT to object detection and dense prediction tasks, many works inherit the multistage design from convolutional networks and highly customized ViT architectures. Behind this design, the goal is to pursue a better trade-off between computational cost and effective aggregation of multiscale global contexts.
However, existing works adopt the multistage architectural design as a black-box solution without a clear understanding of its true benefits.
In this paper, 
we comprehensively study three architecture design choices on ViT -- spatial reduction, doubled channels, and multiscale features -- and demonstrate that a vanilla ViT architecture can fulfill this goal without handcrafting multiscale features, maintaining the original ViT design philosophy.
We further complete a scaling rule to optimize our model's trade-off on accuracy and computation cost / model size.
By leveraging a constant feature resolution and hidden size throughout the encoder blocks, we propose a simple and compact ViT architecture called Universal Vision Transformer (\textbf{UViT}) that achieves strong performance on COCO object detection and instance segmentation benchmark.
Our code will be available at \url{https://github.com/tensorflow/models/tree/master/official/projects/uvit}.
\keywords{Vision Transformer, Self-attention, Object Detection, Instance Segmentation.}
\end{abstract}

\section{Introduction}
\label{sec:intro}

\begin{figure}[t!]
\includegraphics[width=0.8\linewidth]{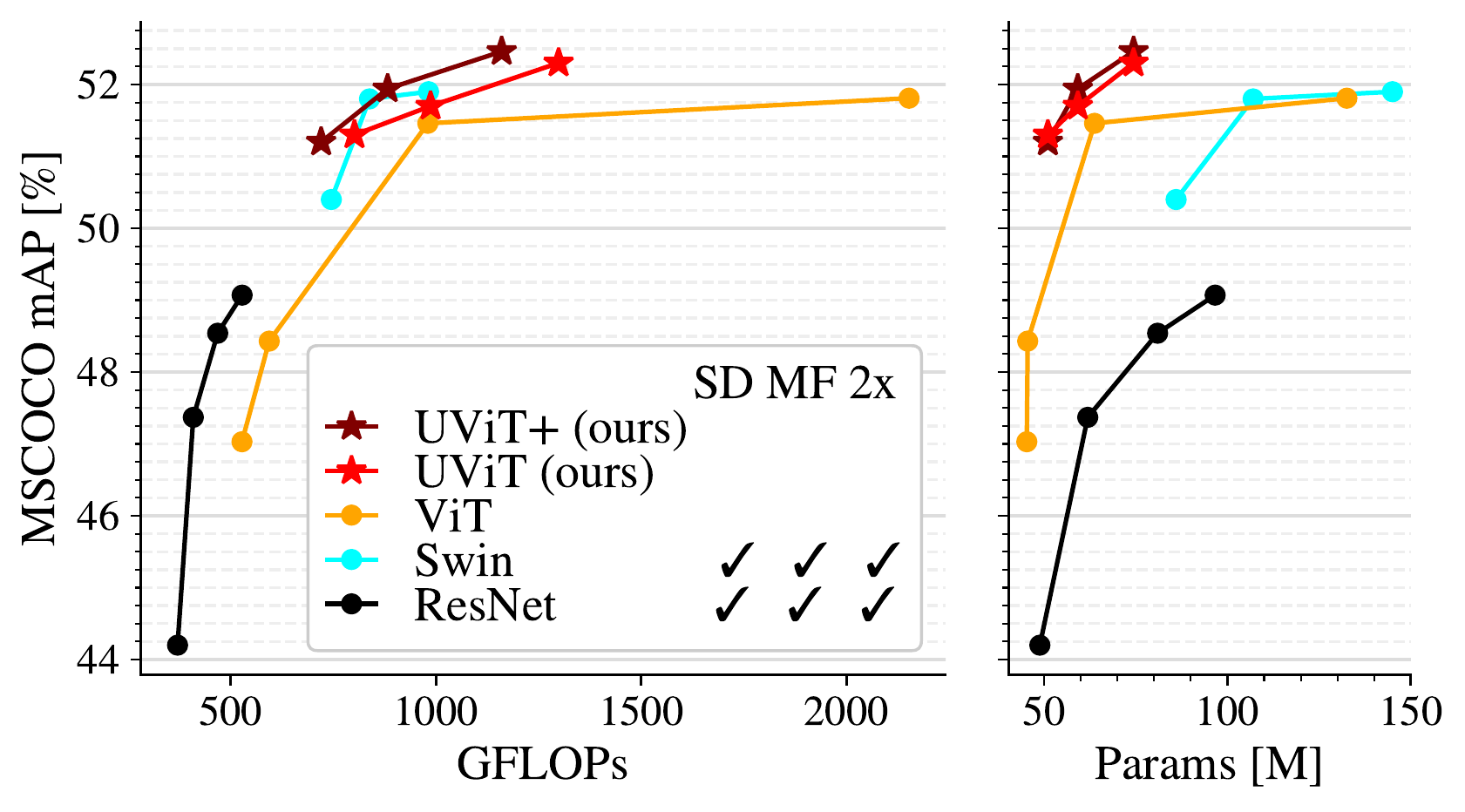}
\centering
\captionsetup{font=small}
\caption{Trade-off between mAP (COCO) and FLOPs (left) / number of parameters (right). We compare our UViT / UViT+ with Swin Transformer~\cite{liu2021swin}, ViT~\cite{zhai2021scaling}, and ResNet (18/50/101/152)~\cite{he2016deep}, all adopting the same standard Cascade Mask RCNN framework~\cite{cai2018cascade}. Our UViT is compact, strong, and simple, avoid using any hierarchical design (``SD'': spatial downsampling, ``MF'': multi-scale features, ``2$\times$'': double channels).} \label{fig:sota_coco}
\vspace{-2em}
\end{figure}

Transformer~\cite{vaswani2017attention}, the de-facto standard architecture for natural language processing (NLP), recently has shown promising results on computer vision tasks.
Vision Transformer (ViT)~\cite{dosovitskiy2020image}, an architecture consisting of a sequence of transformer encoder blocks, has achieved competitive performance compared to convolution neural networks (CNNs)~\cite{he2016deep,simonyan2014very} on the ImageNet classification task~\cite{deng2009imagenet}.

With the success on image classification, recent works extend transformers to more vision tasks, such as object detection~\cite{han2021transformer,huang2021shuffle,liu2021swin}, semantic segmentation~\cite{han2021transformer,huang2021shuffle,liu2021swin,xie2021segformer} and video action recognition~\cite{Arnab2021ViViTAV,Crotts1996VATTColumbiaMS}.
Conventionally, CNN architectures adopt a multistage design with gradually enlarged receptive fields~\cite{he2016deep,simonyan2014very} via spatial reduction or dilated convolutions.
These design choices are also naturally introduced to change vision transformer architectures~\cite{heo2021rethinking,liu2021swin,xie2021segformer}, with two main purposes: 1) \textbf{support of multi-scale features}, since dense vision tasks require the visual understanding of objects of different scales and sizes; 2) \textbf{reduction of computation cost}, for the input images of dense vision tasks are often of high resolutions, and computation complexity of vanilla self-attention is quadratic to the sequence length.
The motivation behind these changes is that, tokens of the original ViT are all of a fixed fine-grained scale throughout the encoder attention blocks, which are not adaptive to dense vision applications, and more importantly incur huge computation/memory overhead.
Despite the success of recent works, it is still unclear if complex design conventions of CNNs are indeed necessary for ViT to achieve competitive performance on vision tasks, and how much benefit comes from each individual design.

In this work, we demonstrate that a vanilla ViT architecture, which we call Universal Vision Transformer (\textbf{UViT}), is sufficient for achieving competitive results on the tasks of object detection and instance segmentation.
Our hope is not ``\textit{to add}'' any special layers to the ViT architecture (thus keeping ViT neat and simple), but instead to choose ``\textit{not to add}'' complex designs that follow CNN structures (multi-scale, double channels, spatial reduction, whose compatibility with attention layers are not thoroughly verified).
In other words, our goal is not to pursue a state-of-the-art performance, but to systematically study the principles in ViT architecture designs.

First, to \textbf{support multiscale features}, instead of re-designing ViT with a multistage fashion~\cite{heo2021rethinking,liu2021swin,xie2021segformer}, our core motivation is that self-attention mechanism naturally encourages the learning of non-local information and makes the feature pyramid no longer necessary for ViTs on dense vision tasks. This leads us to design a simple yet strong UViT architecture: we only leverage constant feature resolution and hidden size throughout the encoder blocks, and extract a single-scale feature map.
Second, to \textbf{reduce the computation cost}, we adopt window splits in attention layers.
We observe that on large input images for detection and instance segmentation, global attentions in early layers are redundant and compact local attentions are both effective and efficient. This motivates us to progressively increase the window sizes as the attention layers become deeper, leading to the drop of self-attention's computation cost with preserved performance.

To support the above two purposes, we systematically study fundamental architecture design choices for UViTs on dense prediction tasks. It is worth noting that, although recent works try to analyze vision transformer's generalization~\cite{naseer2021intriguing}, loss landscapes~\cite{chen2021vision}, and patterns of learned representations~\cite{raghu2021vision,touvron2021going}, they mostly focus on image classification tasks. On dense prediction tasks like object detection and instance segmentation, many transformer models~\cite{liu2021swin,heo2021rethinking,xie2021segformer} directly inherit architecture principles from CNNs, without validating the actual benefit of each individual design choice. In contrast, our simple solution is based on rigorous ablation studies on dense prediction tasks, which is for the first time.
Moreover, we complete a comprehensive study of UViT's compound scaling rule on dense prediction tasks, providing a series of UViT configurations that improve the performance-efficiency trade-off with highly compact architectures (even fewer than 40M parameters on transformer backbone).
Our proposed UViT architectures serve as a simple yet strong baseline on COCO object detection and instance segmentation.

We summarize our contributions as below:
\begin{itemize}[leftmargin=*]
    \item We systematically study the benefits of fundamental architecture designs for ViTs on dense prediction tasks, and propose a simple UViT design that shows strong performance without hand-crafting CNN-like feature pyramid design conventions into transformers.
    \item We discover a new compound scaling rule (depth, width, input size) for UViTs on dense vision tasks. We find a larger input size creates more room for improvement via model scaling, and a moderate depth (number of attention blocks) outperforms shallower or deeper ones.
    \item We reduce the computation cost via only attention windows. We observe that attention's receptive field is limited in early layers and compact local attentions are sufficient, while only deeper layers require global attentions.
    \item Experiments on COCO object detection and instance segmentation demonstrate that our UViT is simple yet a strong baseline for transformers on dense prediction tasks.
\end{itemize}

\section{Related Works}

\subsection{CNN Backbones for Dense Prediction Problems} \label{sec:related_cnns}
CNNs are now mainstream and standard deep network models for dense prediction tasks in computer vision, such as object detection and semantic segmentation. During decades of development, people summarized several high-level and fundamental design conventions: 1) deeper networks for more accurate function approximation~\cite{cohen2016expressive,elbrachter2019deep,eldan2016power,liang2016deep}: ResNet~\cite{he2016deep}, DenseNet~\cite{huang2017densely}; 2) shallow widths in early layers for high feature resolutions, and wider widths in deeper layers for compressed features, which can deliver good performance-efficiency trade-off: Vgg\cite{simonyan2014very}, ResNet\cite{he2016deep}; 3) enlarged receptive fields for learning long-term correlations: dilated convolution (Deeplab series~\cite{chen2017rethinking}), deformable convolutions~\cite{dai2017deformable}; 4) hierarchical feature pyramids for learning across a wide range of object scales: FPN~\cite{lin2017feature}, ASPP~\cite{chen2017rethinking}, HRNet~\cite{wang2020deep}. In short, the motivations behind these successful design solutions fall in two folds: 1) to support the semantic understanding of objects with diverse sizes and scales; 2) to maintain a balanced computation cost under large input sizes. These two motivations, or challenges, also exist in designing our UViT architectures when we are facing dense prediction tasks, for which we provide a comprehensive study in our work (Section~\ref{sec:motivation}).

\subsection{ViT Backbones for Dense Prediction Problems}

The first ViT work~\cite{dosovitskiy2020image} adopted a transformer encoder on coarse non-overlapping image patches for image classification and requires large-scale training datasets (JFT~\cite{sun2017revisiting}, ImageNet-21K~\cite{deng2009imagenet}) for pretraining. DeiT further introduce strong augmentations on both data-level and architecture-level to efficiently train ViT on ImageNet-1k~\cite{deng2009imagenet}.
Beyond image classification, more and more works try to design ViT backbones for dense prediction tasks.
Initially people try to directly learn high-resolution features extracted by ViT backbone via extra interpolation or convolution layers~\cite{beal2020toward,zheng2021rethinking}.
Some works also leverage self-attention operations to replace partial or all convolution layers in CNNs~\cite{hu2019local,ramachandran2019stand,zhao2020exploring}.
More recent trends~\cite{chu2021we,han2021transformer,liu2021swin,wang2021pyramid,xie2021segformer,yuan2021tokens} start following design conventions in CNNs discussed above (Section~\ref{sec:related_cnns}) and customize ViT architectures to be CNN-like: tokens are progressively merged to downsample the feature resolutions with reduced computation cost, along with increased embedding sizes. Multi-scale feature maps are also collected from the ViT backbone. These ViT works can successfully achieve strong or state-of-the-art performance on object detection or semantic segmentation, but the architecture is again highly customized for vision problems and lose the potential for multi-modal learning in the future. More importantly, those CNN-like design conventions are directly inherited into ViTs without a clear understanding of each individual benefit, leading to empirical black-box designs.
In contrast, the simple and neat solution we will provide is motivated by a complete study on ViT's architecture preference on dense prediction tasks (Section~\ref{sec:motivation} and~\ref{sec:main_solution}).

\subsection{Inductive Bias of ViT Architecture}

Since the architecture of vision transformers is still in its infant stage, there are few works that systematically study principles in ViT's model design and scaling rule.
Initially, people leverage coarse tokenizations, constant feature resolution, and constant hidden size~\cite{dosovitskiy2020image,touvron2020training}, while recently fine-grained tokens, spatial downsampling, and doubled channels are also becoming popular in ViT design~\cite{liu2021swin,zhou2021deepvit}. They all achieve good performance, calling for an organized study on the benefits of different fundamental designs.
In addition, different learning behaviors of self-attentions (compared with CNNs) make the scaling law of ViTs highly unclear.
Recent works~\cite{raghu2021vision} revealed ViT generates more uniform receptive fields across layers, enabling the aggregation of global information in early layers.
This is contradictory to CNNs which require deeper layers to help the learning of visual global information~\cite{chen2018encoder}.
Attention scores of ViTs are also found to gradually become indistinguishable as the encoder goes deeper, leading to identical and redundant feature maps, and plateaued performance~\cite{zhou2021deepvit}.
These observations all indicate that previously discovered design conventions and scaling laws for CNNs~\cite{he2016deep,tan2019efficientnet} may not be suitable for ViTs, thus calling for comprehensive studies on the new inductive bias of ViT's architecture on dense prediction tasks.

\section{Methods}

Our work targets designing a simple ViT model for dense prediction tasks, and trying to avoid hand-crafted customization on architectures.
We will first explain our motivations with comprehensive ablation studies on individual design benefits in Section~\ref{sec:motivation}, and then elaborate the discovered principles of our UViT designs in Section~\ref{sec:main_solution}.

\subsection{Is a Simple ViT Design All You Need?} \label{sec:motivation}

As discussed in Section~\ref{sec:intro}, traditionally CNNs leverage resolution downsampling, doubled channels, and hierarchical pyramid structures, to support both multi-scale features and reduction of computation cost~\cite{chen2017rethinking,he2016deep,lin2017feature}.
Although recent trends in designing ViTs also inherit these techniques, it is still unclear whether they are still beneficial to ViTs.
Meanwhile, ViT~\cite{dosovitskiy2020image}, DeiT~\cite{touvron2020training}, and T2T-ViT~\cite{yuan2021tokens} demonstrate that, at least for image classification, a constant feature resolution and hidden size can also achieve competitive performance. Without spatial downsampling, the computation cost of self-attention blocks can also be reduced by using attention window splits~\cite{beltagy2020longformer,liu2021swin}, i.e., to limit the spatial range of the query and the key when we calculate the dot product attention scores.
To better understand each individual technique and to systematically study the principles in ViT architecture designs, we provide a comprehensive study on the contributions of CNN-like design conventions to ViTs on dense prediction tasks.

\paragraph{Implementations:} We conduct this study on the object detection task on COCO 2017 dataset. We leverage the standard Cascade Mask-RCNN detection framework~\cite{cai2018cascade,he2017mask}, with a fixed input size as $640\times640$. All detection models are fine-tuned from an ImageNet pretrained initialization. More details can be found in Section~\ref{sec:implementations}.

\paragraph{Settings:} We systematically study the benefit of spatial downsampling, multi-scale features, and doubled channels to the object detection performance of ViT. We start from a baseline ViT architecture close to the S16 model proposed in~\cite{dosovitskiy2020image}, which has 18 attention blocks, a hidden size of 384, and six heads for each self-attention layer. The first linear project layer embeds images into $\frac{1}{8}$-scale patches, i.e., the input feature resolution to the transformer encoder is $\frac{1}{8}$. The attention blocks will be grouped into three stages for one or a combination of two or three purposes below:
\begin{itemize}[leftmargin=*]
    \item Spatial downsampling: tokens will be merged between two consecutive stages to downsample the feature resolution. If the channel number is also doubled between stages, the tokens will be merged by a learned convolution with a stride as 2; otherwise, tokens will be merged by a 2D bi-linear interpolation.
    \item Multi-scale features: after each stage, features of a specific resolution will be output and fed into the detection FPN head. Multi-scale features of three target resolutions ($\frac{1}{8}, \frac{1}{16}, \frac{1}{32}$) will be collected from the encoder from early to deep attention layers.
    \item Doubled channels: after each of the first two stages, the token's hidden size will be doubled via a linear projection.
\end{itemize}

\begin{figure}[t]
\includegraphics[width=0.7\linewidth]{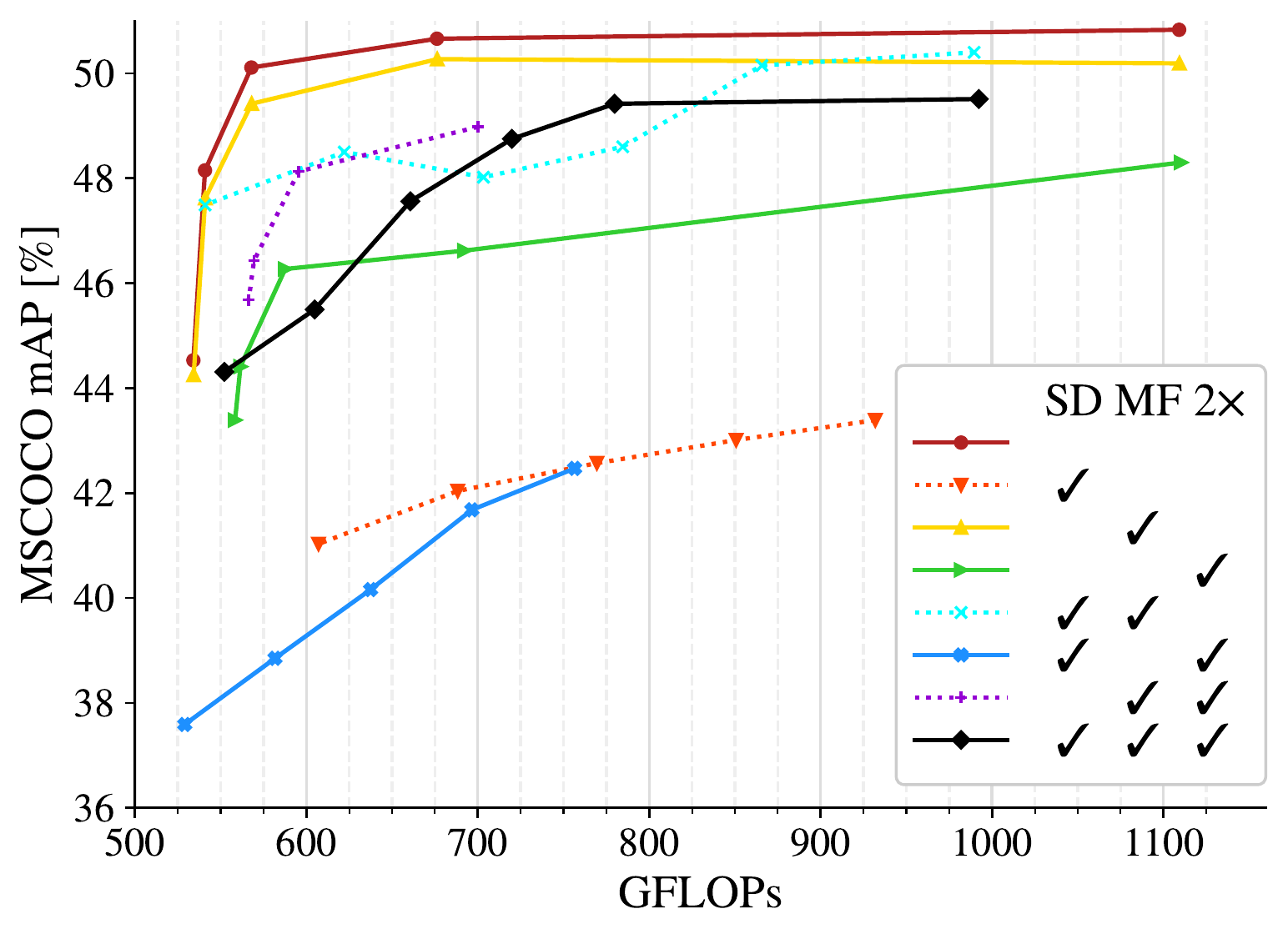}
\centering
\caption{The benefits of various commonly used CNN-inspired design changes to ViT: spatial downsampling (``SD''), multi-scale features (``MF''), and doubled channels (``$2\times$'').
With controlled number of parameters (72M) and input size (640$\times$640), not using any of these designs and sticking to the original ViT model~\cite{dosovitskiy2020image} performs the best in a wide range of FLOPs we explore.} \label{fig:ms_ablation}
\vspace{-1em}
\end{figure}

We study all combinations of the above three techniques, i.e. eight settings in total, and show the results in Figure~\ref{fig:ms_ablation}. Note that each dot in Figure~\ref{fig:ms_ablation} indicates an individually designed and trained model. All models are of around 72 million parameters, and are trained and evaluated under the same $640\times 640$ input size. Therefore, for vertically aligned dots, they share the same FLOPs, number of parameters, and input size, thus being fairly comparable. We control the FLOPs (x-axis) by changing the depths or attention windows allocated to different stages, see our supplement for more architecture details.

\paragraph{Observations}
\begin{itemize}[leftmargin=*]
    \item Spatial Downsampling (``SD'') does not seem to be beneficial. Our hypothesis is that, under the same FLOPs constraint, the self-attention layers already provide global features, and do not need to downsample the features to enlarge the receptive field.
    \item Multi-scale Features (``MF'') can mitigate the poor performance from downsampling by leveraging early high-resolution features (``SD$+$MF''). However, the vanilla setting still outperforms this combination.
    We hypothesize that high-resolution features are extracted too early in the encoder; in contrast, tokens in vanilla ViTs are able to learn fine-grained details throughout the encoder blocks.
    \item Doubled channels (``$2\times$'') plus multi-scale features (``MF'') may potentially seem competitive. However, ViT does not show strong inductive bias on ``deeper compressed features with more embedding dimension''. This observation is also aligned with findings in~\cite{raghu2021vision} that ViTs have highly similar representations throughout the model, indicating that we should not sacrifice embedding dimensions of early layers to compensate for deeper layers.
\end{itemize}

In summary, we did not find strong benefits by adopting CNN-like design conventions. Instead, A simple architecture solution of a constant feature resolution and hidden size could be a strong ViT baseline.

\vspace{0.5em}
\subsection{UViT: a Simple Yet Effective Solution} \label{sec:main_solution}

\begin{figure*}[t!]
\includegraphics[scale=0.43]{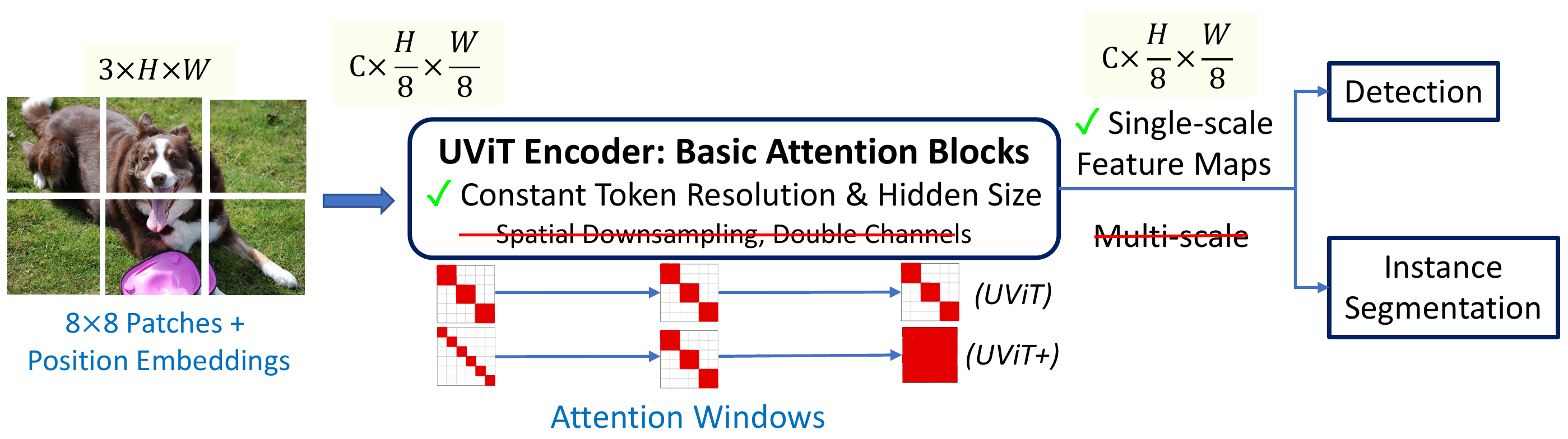}
\centering
\captionsetup{font=small}
\caption{We keep the architecture of our UViT neat: image patches (plus position embeddings) are processed by a stack of vanilla attention blocks with a constant resolution and hidden size. Single-scale feature maps as outputs are fed into head modules for detection or segmentation tasks. Constant (UViT, Section~\ref{sec:scaling}) or progressive (UViT+, Section~\ref{sec:window_strategy}) attention windows are introduced to reduce the computation cost. We demonstrate that this simple architecture is strong, without introducing design overhead from hierarchical spatial downsampling, doubled channels, and multi-scale feature pyramids.} \label{fig:overview}
\end{figure*}

Based on our study in Section~\ref{sec:motivation}, we are motivated to simplify the ViT design for dense prediction tasks and provide a neat solution, as illustrated in Figure~\ref{fig:overview}.
Taking $8\times 8$ patches of input images, we learn the representation by using a constant token resolution of $\frac{1}{8}$ scale (the number of tokens remains the same) and a constant hidden size (the channel number will not be increased). 
A single-scale feature map will be fed into a detection or segmentation head.
Meanwhile, attention windows~\cite{beltagy2020longformer} will be leveraged to reduce the computation cost.

The motivation behind our UViT design is not ``\textit{to add}'' any layers to the ViT architecture, but instead to choose ``\textit{not to add}'' complex designs.
Our detailed study on input/model scaling will demonstrate that, a vanilla ViT architecture plus a better depth-width trade-off can achieve a high performance.

Though being simple, still we have two core questions to be determined in our design: (1) How to balance the UViT's depth, width, and input size to achieve the best performance-efficiency trade-off? (Section~\ref{sec:scaling}) (2) Which attention window strategy can effectively save the computation cost without sacrificing the performance? (Section~\ref{sec:attn_window})

\subsubsection{A Compound Scaling Rule of UViTs\\} \label{sec:scaling} 

\vspace{0.5em}

\begin{figure*}[t!]
\centering
\includegraphics[scale=0.32]{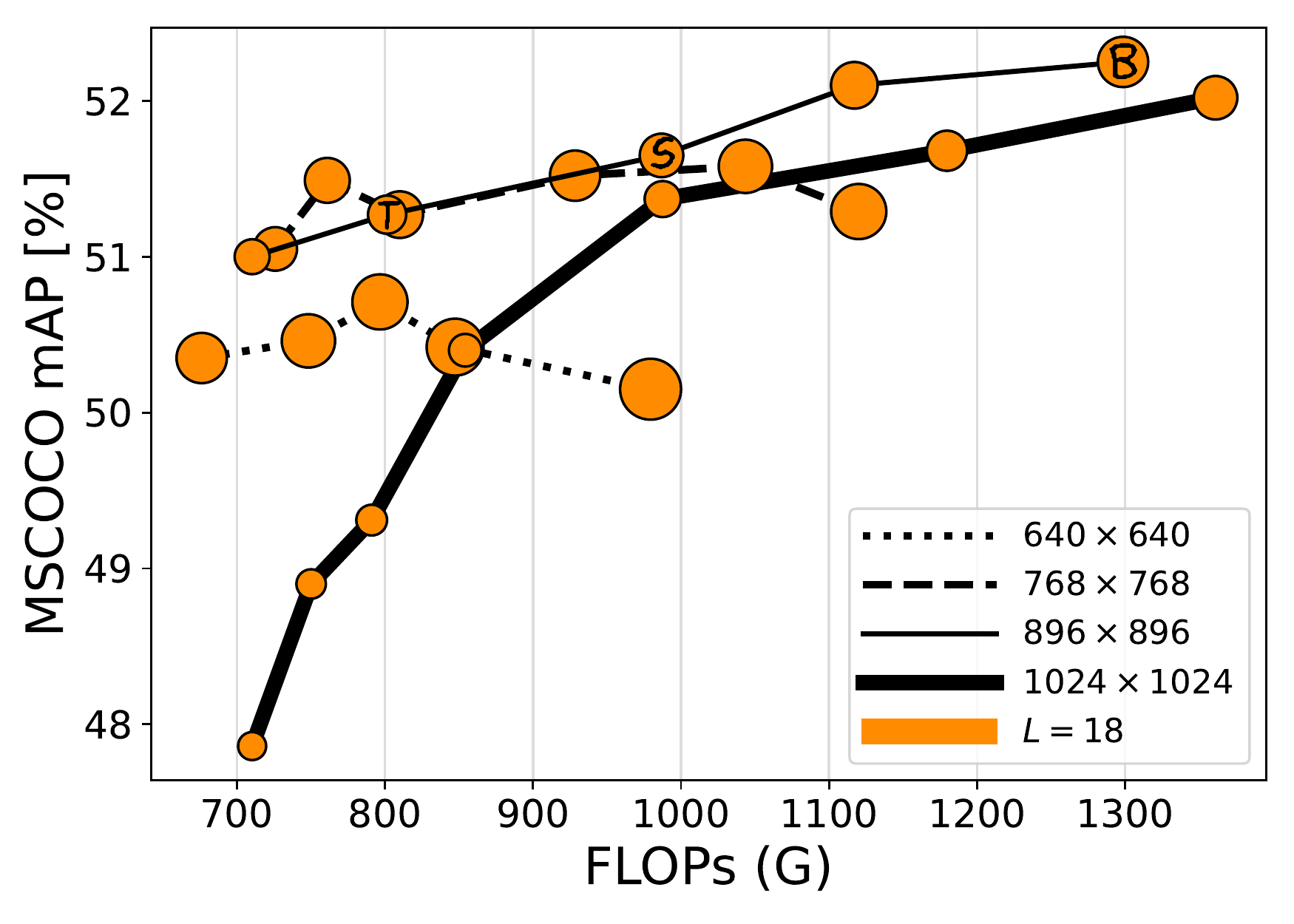}
\hspace{1em}
\includegraphics[scale=0.32]{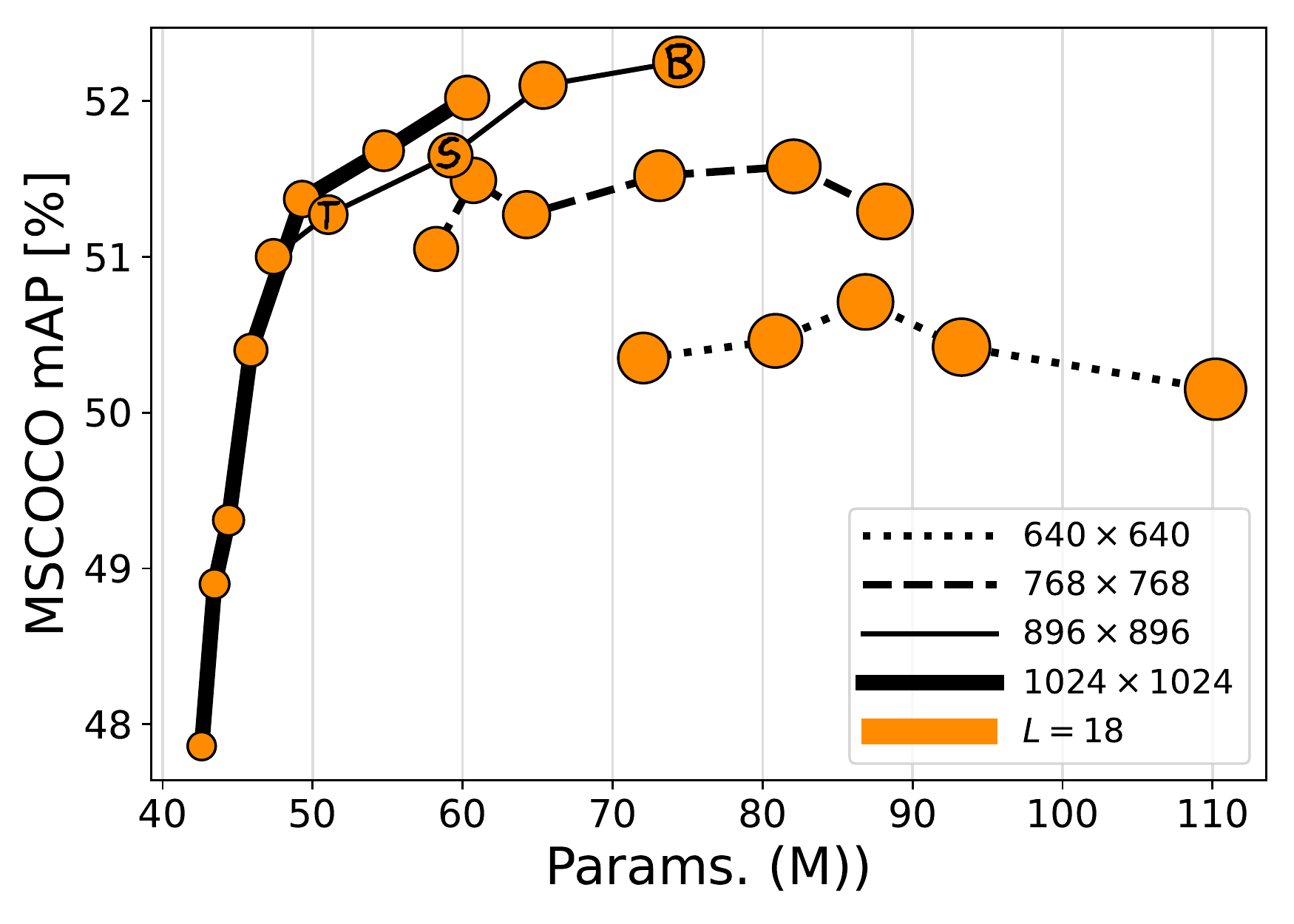}
\centering
\captionsetup{font=small}
\caption{\textbf{Input} scaling rule for UViT on COCO object detection. Given a fixed depth, an \textbf{input size of $\bm{896}\times \bm{896}$} (thin solid line) leaves more room for model scaling (by increasing the width) and is slightly better than $1024\times 1024$ (thick solid line); and $640\times 640$ (dashed line) or $768\times 768$ (dotted line) are of worse performance-efficiency trade-off. Black capital letters ``\textit{T}'', ``\textit{S}'', and ``\textit{B}'' annotate three final depth/width configurations of UViT variants we will propose (Table~\ref{table:architecture}). Different sizes of markers represent the hidden sizes (widths).} \label{fig:scaling_rule_input_size}
\vspace{-1.5em}
\end{figure*}

\begin{figure*}[t!]
\centering
\includegraphics[scale=0.33]{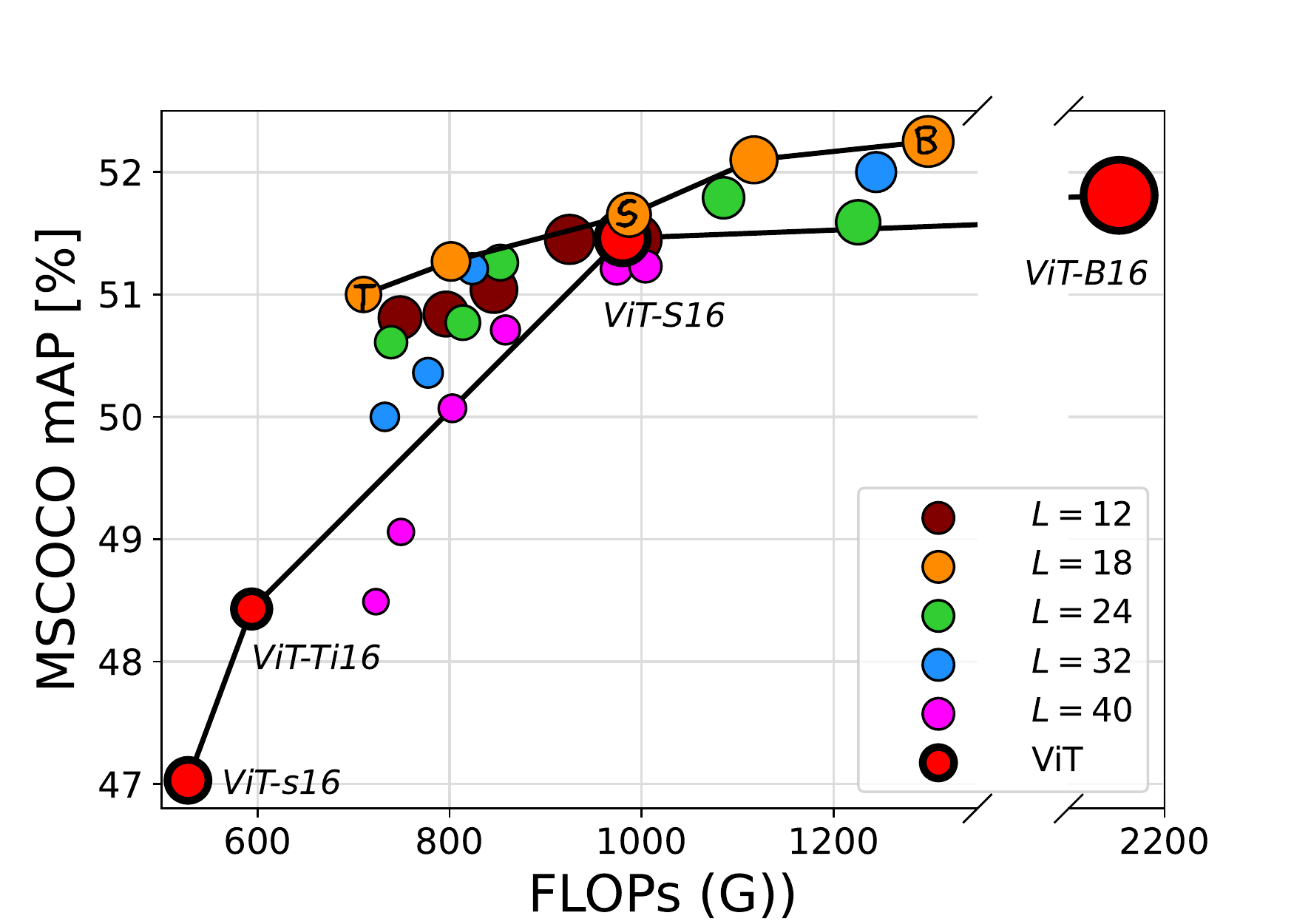}
\includegraphics[scale=0.33]{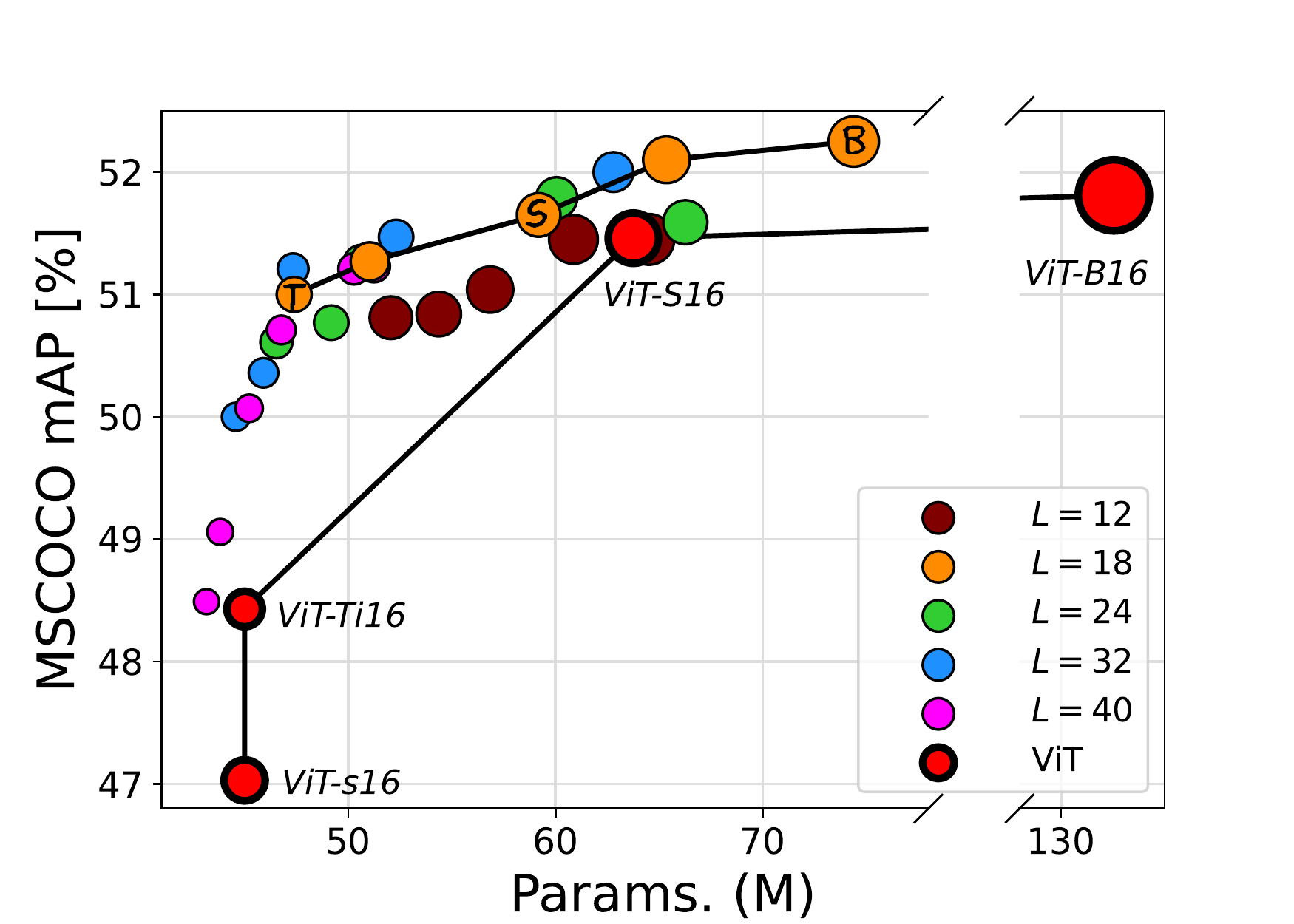}
\centering
\captionsetup{font=small}
\caption{\textbf{Model} scaling rule for UViT on COCO object detection. \textbf{18 attention blocks} (\textcolor{orange}{orange}), which provide a balanced trade-off between depth and width, performs better than shallower or deeper UViTs. Black capital letters ``\textit{T}'', ``\textit{S}'', and ``\textit{B}'' annotate three final depth/width configurations of UViT variants we will propose (Table~\ref{table:architecture}). Different sizes of markers represent the hidden sizes (widths).} \label{fig:scaling_rule_model}
\vspace{-1.5em}
\end{figure*}

Previous works studied compound scaling rules for CNNs~\cite{tan2019efficientnet} and ViTs~\cite{zhai2021scaling} on image classification. However, few works studied the scaling of ViTs on dense prediction tasks.
To achieve the best performance-efficiency trade-off, we systematically study the compounding scaling of UViTs on three dimensions: input size, depth, and widths. We show our results in Figure~\ref{fig:scaling_rule_input_size} and Figure~\ref{fig:scaling_rule_model}\footnote{This compound scaling rule is studied in Section~\ref{sec:scaling} before we study the attention window strategy in Section~\ref{sec:attn_window}. Thus for all models in Figure~\ref{fig:scaling_rule_input_size} and Figure~\ref{fig:scaling_rule_model} we adopt the window scale as $\frac{1}{2}$, for fair comparisons.}.
For all models (circle markers), we first train them on ImageNet-1k, then fine-tune them on the COCO detection task.
\begin{itemize}[leftmargin=*]
    \item Depth (number of attention blocks): we study different UViT models of depths selected from $\{12, 18, 24, 32, 40\}$.
    \item Input size: we study three levels of input sizes: $640\times 640$, $768\times 768$, $896\times 896$, and $1024\times 1024$.
    \item Width (i.e. hidden size, or output dimension of attention blocks): we will tune the width to further control different model sizes and computation costs to make different scaling rules fairly comparable.
\end{itemize}
\paragraph{Observations}
\begin{itemize}[leftmargin=*]
    \item In general, UViT can achieve a strong mAP with a moderate computation cost (FLOPs) and a highly compact number of parameters (even fewer than 70M including the Cascaded FPN head).
    \item For input sizes (Figure~\ref{fig:scaling_rule_input_size} by different line styles): large inputs generally create more room for models to further scale up. Across a wide range of model parameters and FLOPs, we find that the scaling under an $896\times 896$ input size constantly outperforms smaller input sizes (which lead to severe model overfitting), and is also better than $1024\times 1024$ in a comparable FLOPs range.
    \item For the model depths (Figure~\ref{fig:scaling_rule_model}), different depths in colors): we find that considering both FLOPs and the number of parameters, 18 blocks achieve better performance than 12/24/32/40 blocks. This indicates UViT needs a balanced trade-off between depth and width, instead of sacrificing depth for more width (e.g. 12 blocks) or sacrificing width for more depth (e.g. 40 blocks).
\end{itemize}

In summary, based on our final compound scaling rule, we propose our basic version of UViT as 18 attention blocks under $896\times 896$ input size.
See our supplement for more architecture details.

\subsubsection{Attention Windows: a Progressive Strategy\\} \label{sec:attn_window}

In this section, we will show that a progressive attention window strategy can reduce UViT's computation cost while still preserving or even increasing the performance. 

\paragraph{Early attentions are local operators.}

Originally, self-attention~\cite{vaswani2017attention} is a global operation: unlike convolution layers that share weights to local regions, any pair of tokens in the sequence will contribute to the feature aggregation, thus collecting global information to each token. In practice, however, self-attention in different layers may still have biases in regions they prefer to focus on.

To validate this assumption, we select a pretrained ViT-B16 model~\cite{dosovitskiy2020image}, and calculate the relative receptive field of each self-attention layer on COCO. Given a sequence feature of length $L$ and the attention score $\bm{s}$ (after softmax) from a specific head, the relative receptive field $r$ is defined as:
\begin{equation}
    r = \frac{1}{L} \sum_{i=1}^L \frac{\sum_{j = 1}^L \bm{s}_{i,j} |i - j|}{\max(i, L - i)}, i,j = 1, \cdots, L,
\end{equation}
where $\sum_{j=1}^L \bm{s}_{i,j} = 1$ for $j = 1, \cdots, L$. This relative receptive filed takes into consideration the token's position and the furthest possible location a token can aggregate, and indicates the spatial focus of the self-attention layer.

\begin{figure}[h!]
\vspace{-1em}
\includegraphics[scale=0.45]{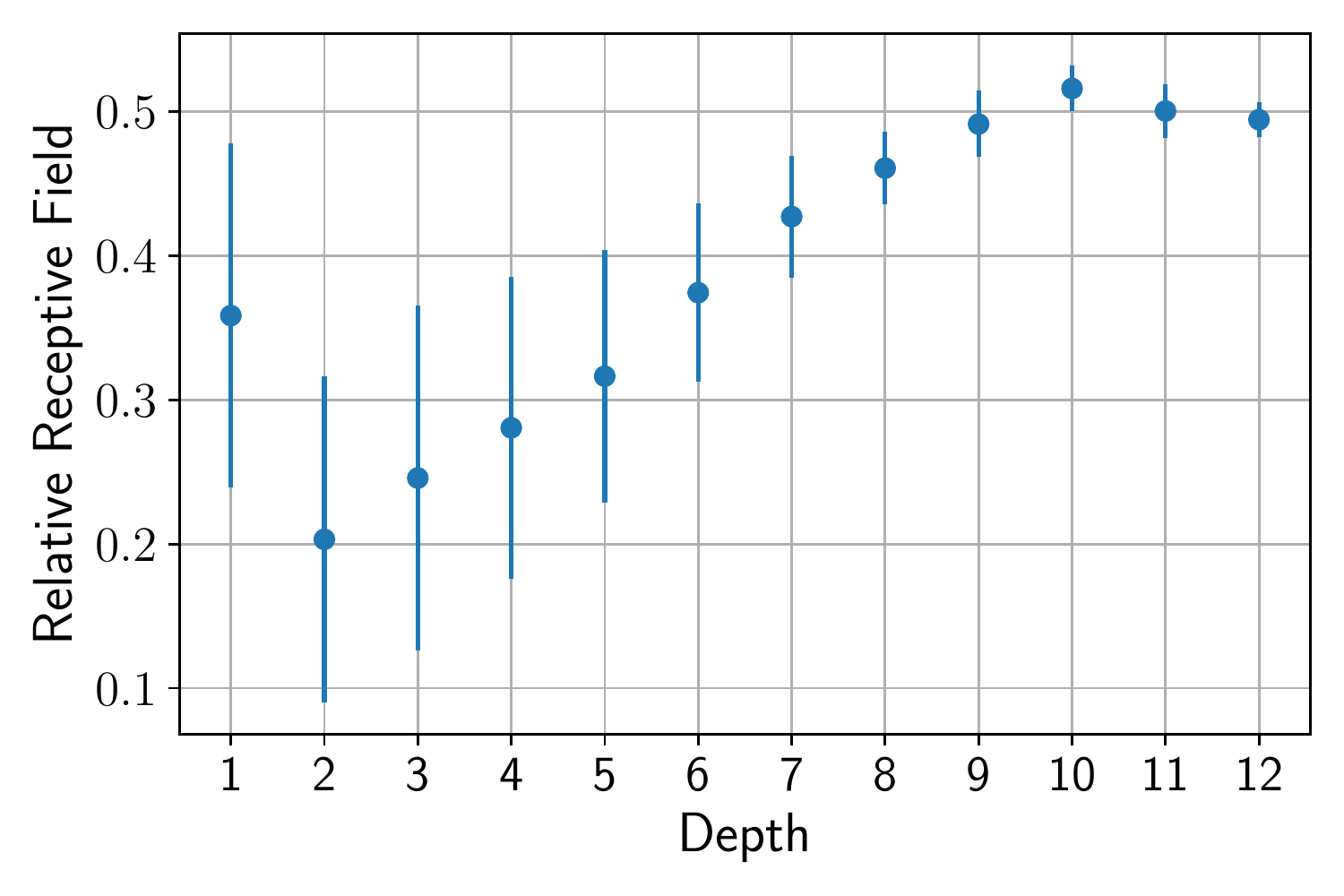}
\vspace{-1em}
\centering
\captionsetup{font=small}
\caption{Relative attention's receptive field of a ImageNet pretrained ViT-B16~\cite{dosovitskiy2020image} along depth (indices of attention blocks), on the COCO dataset. Error bars are standard deviations across different attention heads.} \label{fig:receptive_field} \label{fig:search_space}
\vspace{-2em}
\end{figure}

We collect the averages and standard deviations across different attention heads.
As shown in Figure~\ref{fig:receptive_field}, we can see that tokens in early attention layers, although having the potential to aggregate long-range features, weight more on their neighbor tokens and thus act like a ``local operator''. As the attention layers stack deeper, the receptive field increases, transiting the self-attention to a global operation. This inspires us that, if we explicitly limit the attention range of early layers, we may save the computation cost but still preserve the capability of self-attentions.

\paragraph{Attention window improves UViT's efficiency.} \label{sec:window_strategy}

Motivated by Figure~\ref{fig:receptive_field}, we want to study the most effective attention window strategy. Specifically, we need principled answers to two design questions:
\begin{enumerate}
    \item[1)] \textit{How small a window size can early attention layers endure?} To answer this question, we start the attention blocks with square windows with different small scales: $\{\frac{1}{16}, \frac{1}{8}, \frac{1}{4}\}$ of height or width a sequence's 2D shape\footnote{For example, if the input sequence has $(896/8)\times(896/8) = 112\times 112$ tokens, a window of scale $\frac{1}{16}$ will contain $7\times 7 = 49$ elements. Similar ideas for $\frac{1}{8}$ and $\frac{1}{4}$.}.
    \item[2)] \textit{Do deeper layers require global attention, or some local attentions are also sufficient?} To compare with global attentions (window size as $1$), we will also try small attention windows (window size of $\frac{1}{2}$ scale) in deeper layers.
\end{enumerate}

To represent an attention window strategy that ``progressively increases window scales from $\frac{1}{4}$ to $\frac{1}{2}$ to $1$'', we use a simple annotation ``$[4^{-1}]\times 14 \rightarrow [2^{-1}]\times 2 \rightarrow [1]\times 2$'', indicating that there are 14 attention blocks assigned with $\frac{1}{4}$-scale windows, then two attention blocks assigned with $\frac{1}{2}$-scale windows, and finally two attention blocks assigned with $1$-scale windows.
When comparing different window strategies, we make sure all strategies have the same number of parameters and share the similar computation cost for fair comparisons. We also include four more baselines of constant attention window scale across all attention blocks: global attention, and also windows of $\frac{1}{4} \sim \frac{1}{2}$ scale.
We show our results in Table~\ref{table:window_ablation}, and summarize observations below:\vspace{0.5em}

\begin{table*}[h!]
\small
\caption{Over-shrank window sizes in early layers are harmful, and global attention windows in deep layers are vital to the final performance. Fractions in brackets indicate attention window scales (relative to sequence feature sizes), and the multiplier indicates the number of attention blocks allocated to an attention window scale (18 blocks in total). Standard Deviations of three random runs are shown in parentheses.}
\vspace{0.5em}
\centering
\resizebox{0.95\textwidth}{!}{
\setlength{\tabcolsep}{1.5pt}
\begin{tabular}{c  c  cc}
\toprule
$[\mathrm{window\_scale}] \times \#\mathrm{layers}$ & GFLOPs & \text{AP$_\text{val}$} & Img/s \\ \midrule
$[1]\times 18$ & 2961.9 & 52.4 (0.09) & 3.5 \\
$[2^{-1}]\times 18$ & 1298.7 & 52.3 (0.17) & 10.5 \\ \midrule
$[16^{-1}]\times 4 \shortrightarrow [8^{-1}]\times 4 \shortrightarrow [4^{-1}]\times 4 \shortrightarrow [2^{-1}]\times 4 \shortrightarrow [1]\times 2$ & 1154.3 & 52.0 (0.15) & 11.5 \\
$[8^{-1}]\times 9 \rightarrow [4^{-1}]\times 4 \rightarrow [2^{-1}]\times 3 \rightarrow [1]\times 2$ & 1131.2 & 52.2 (0.21) & 12.7 \\
$[4^{-1}]\times 14 \rightarrow [2^{-1}]\times 2 \rightarrow [1]\times 2$ & 1160.1 & 52.5 (0.11) & 12.3 \\
$[4^{-1}]\times 6 \rightarrow [2^{-1}]\times 12$ & 1160.1 & 52.2 (0.12) & 12.5 \\ \bottomrule
\end{tabular}
}
\label{table:window_ablation}
\end{table*}

\begin{itemize}
    \item With smaller constant window scale ($\frac{1}{2}/\frac{1}{3}/\frac{1}{4}$), we save more computation cost with slight sacrifice in mAP.
    \item Adopting constant global attentions throughout the whole encoder blocks (window size as $1$, first row) is largely redundant, which contributes marginal benefits but suffers from a huge computation cost.
    \item Early attentions can use smaller windows like $\frac{1}{4}$-scale, but over-shrank window sizes ($\frac{1}{16}, \frac{1}{8}$) can impair the capability of self-attentions (3rd, 4th rows).
    \item Deeper layers still require global attentions to preserve the final performance (last two rows).
    \item A properly designed window strategy (5th row) can outperform vanilla solutions (1st, 2nd row) with a reduced computation cost.
\end{itemize}

In conclusion, we set the window scale of our basic version (UViT, Section~\ref{sec:scaling}) as constant $2^{-1}$, and proposed an improved version of our model, dubbed ``UViT+'' with the attention window strategy adopted as ``$[4^{-1}]\times 14 \rightarrow [2^{-1}]\times 2 \rightarrow [1]\times 2$''.

\section{Final Results}

We conduct our experiments on COCO~\cite{lin2014microsoft} object detection and instance segmentation to show our final performance.

\subsection{Implementations} \label{sec:implementations}
We implement our model and training in TensorFlow and Keras. Experiments are conducted on TPUs. Before fine-tuning on object detection or instance segmentation, we follow the DeiT~\cite{touvron2020training} training settings to pretrain our UViTs on ImageNet-1k with a $224\times 224$ input size and a batch size of 1024.
We follow the convention in~\cite{dosovitskiy2020image}: during ImageNet pretraining the kernel size of the first linear projection layer is $16\times 16$. During fine-tuning, we will use a more fine-grained $8\times 8$ patch size for the dense sampling purpose. The kernel weight of the first linear project layer will be interpolated from $16\times 16$ to $8\times 8$, and the position embedding will also be elongated by interpolation.
We also report the throughput (``Img/s''), which measures the latency of UViTs by feeding one image per TPU core.

\subsection{Architectures and ImageNet Pretraining}

\begin{table}[h]
\captionsetup{font=small}
\centering
\small
\captionsetup{font=small}
\caption{Architecture variants of our UViT with ImageNet~\cite{deng2009imagenet} Pretraining Performance.}
\vspace{0.5em}
\resizebox{0.8\textwidth}{!}{
\begin{tabular}{ccccccc}
\toprule
Name & Depth & Hidden Size & Params. (M) & GFLOPs & Top-1 & Img/s \\ \midrule
UViT-T & 18 & 222 & 13.5 & 2.5 & 76.0\% & 170.2 \\
UViT-S & 18 & 288 & 21.7 & 4.0 & 78.9\% & 145.4 \\
UViT-B & 18 & 384 & 32.8 & 6.9 & 81.3\% & 134.2\\ \bottomrule
\end{tabular}}\label{table:architecture}
\end{table}

We propose three variants of our UViT variants.
The architecture configurations of our model variants are listed in Table~\ref{table:architecture}, and are also annotated in Figure~\ref{fig:scaling_rule_input_size} and Figure~\ref{fig:scaling_rule_model} (``T'', ``S'', ``B'' in black).
The number of heads is fixed as six, and the expansion ratio of each FFN (feed-forward network) layer if fixed as four in all experiments. As discussed in Section~\ref{sec:attn_window}, the attention window strategy will be ``$[4^{-1}]\times 14 \rightarrow [2^{-1}]\times 2 \rightarrow [1]\times 2$''.

\subsection{COCO detection \& instance segmentation}

\begin{table*}[h!]\centering
\small
\captionsetup{font=small}
\caption{Two-stage object detection and instance segmentation results on COCO 2017.
We compare employing different backbones with Cascade Mask R-CNN on single model without test-time augmentation. UViT sets a constant window scale as $2^{-1}$, and UViT+ adopts the attention window strategy as ``$[4^{-1}]\times 14 \rightarrow [2^{-1}]\times 2 \rightarrow [1]\times 2$''. We also reproduced the performance of ResNet under the same settings.}
\vspace{0.5em}
\resizebox{1\textwidth}{!}{
\setlength{\tabcolsep}{1.5pt}
\begin{tabular}{c  cccccc}
\toprule
   \multicolumn{1}{c}{Backbone} & \hspace{3mm}Resolution\hspace{5mm} &\text{GFLOPs} & \text{Params. (M)}
 & \hspace{2mm}\text{AP$_\text{val}$}  \hspace{2mm} & \text{AP$^{\text{mask}}_\text{val}$} & Img/s\\ \midrule
    ResNet-18 & 896$\times$896 & 370.4 & 48.9 & 44.2 & 38.5 & - \\
    ResNet-50 & 896$\times$896 & 408.8 & 61.9 & 47.4 & 40.8 & - \\
    Swin-T~\cite{liu2021swin} & 480$\sim$800$\times$1333 & 745 & 86 & 50.5 & 43.7 & 15.3 \\
    Shuffle-T~\cite{huang2021shuffle} & 480$\sim$800$\times$1333 & 746 & 86 & 50.8 & 44.1 & - \\
    UViT-T (ours) & 896$\times$896 & 801.4 & 51.0 & \textbf{51.3} & 43.6 & 11.8 \\
    UViT-T+ (ours) & 896$\times$896 & 720.2 & 51.0 & \textbf{51.2} & \textbf{43.9} & 14.2 \\
    \midrule
    ResNet-101 & 896$\times$896 & 468.2 & 81.0 & 48.5 & 41.8 & - \\
    Swin-S~\cite{liu2021swin} & 480$\sim$800$\times$1333 & 838 & 107 & 51.8 & 44.7 & 12.0 \\
    Shuffle-S~\cite{huang2021shuffle} & 480$\sim$800$\times$1333 & 844 & 107 & \textbf{51.9} & \textbf{44.9} & - \\
    UViT-S (ours) & 896$\times$896 & 986.8 & 59.2 & 51.7 & 44.1 & 11.1 \\
    UViT-S+ (ours) & 896$\times$896 & 882.2 & 59.2 & \textbf{51.9} & 44.5 & 12.5 \\
    \midrule
    ResNet-152 & 896$\times$896 & 527.7 & 96.7 & 49.1 & 42.1 & - \\
    Swin-B~\cite{liu2021swin} & 480$\sim$800$\times$1333 & 982 & 145 & 51.9 & 45 & 11.6 \\
    Shuffle-B~\cite{huang2021shuffle} & 480$\sim$800$\times$1333 & 989 & 145 & 52.2 & 45.3 & - \\
    GCNet~\cite{cao2020global} & - & 1041 & - & 51.8 & 44.7 & - \\
    UViT-B (ours) & 896$\times$896 & 1298.7 & 74.4 & 52.3 & 44.3 & 10.5 \\
    UViT-B+ (ours) & 896$\times$896 & 1160.1 & 74.4 & \textbf{52.5} & 44.8 & 12.3 \\
    \begin{tabular}{@{}c@{}}UViT-B+ (ours) \\ w/ self-training\end{tabular} & 896$\times$896 & 1160.1 & 74.4 & \textbf{53.9} & \textbf{46.1} & 12.3 \\ \bottomrule
\end{tabular}
}
\label{table:detection}
\end{table*}

\paragraph{Settings} Object detection experiments are conducted on COCO 2017~\cite{lin2014microsoft}, which contains 118K training and 5K validation images. We consider the popular Cascade Mask-RCNN detection framework~\cite{cai2018cascade,he2017mask}, and leverage multi-scale training~\cite{sun2021sparse,carion2020end} (resizing the input to $896\times 896$), AdamW optimizer~\cite{loshchilov2017decoupled} (with an initial learning rate as $3\times 10^{-3}$), weight decay as $1\times 10^{-4}$, and a batch size of 256. Similar above, the thoughput (``Img/s'') measures the latency of UViTs with one COCO image per TPU core.

From Table~\ref{table:detection} we can see that on different levels of model variants, our UViTs are highly compact. Compared with both CNNs and other ViT works, our UViT achieves strong results with much better efficiency: with similar GFLOPs, UViT uses a much fewer number of parameters (at least \textbf{44.9\%} parameter reduction compared with Swin~\cite{liu2021swin}). To make this comparison clean, we did not adopt any system-level techniques~\cite{liu2021swin} to boost the performance\footnote{As we adopt the popular Cascade Mask-RCNN detection framework~\cite{cai2018cascade,he2017mask}, some previous detection works~\cite{chen2020reppoints,sun2021sparse}
may not be directly compared.
}. As we did not leverage any CNN-like hierarchical pyramid structures, the results of our simple and neat solution suggest that, the original design philosophy of ViT~\cite{dosovitskiy2020image} is a strong baseline without any hand-crafted architecture customization. We also show the mAP-efficiency trade-off curve in Figure~\ref{fig:sota_coco}.
Besides, we also adopt our UViT-B backbone with the Mask-RCNN~\cite{he2017mask} framework, and achieve 50.5 \text{AP$_\text{val}$} with 1026.1 GFLOPs.

Additionally, we adopt self-training on top of our largest model (UViT-B) to evaluate the performance gain by leveraging unlabeled data, similar as~\cite{zoph2020rethinking}. We use ImageNet-1K without labels as the unlabeled set, and a pretrained UViT-B model as the teacher model to generate pseudo-labels. All predicted boxes with confidence scores larger than 0.5 are kept, together with their corresponding masks. For UViT-B with self-training, the student model is initialized from the same weights of the teacher model. The ratio of labeled data to pseudo-labeled data is 1:1 in each batch. Apart from increasing training steps by $2\times$ for each epoch, all other hyperparameters remain unchanged. We can see from the last row in Table~\ref{table:detection} that self-training significantly improves box AP and mask AP by 1.4\% and 1.3\%, respectively.

\section{Conclusion}
We present a simple, single-scale vision transformer backbone that can serve as a strong baseline for object detection and semantic segmentation.
Our novelty is not ``to add'' any special layers to ViT, but instead to choose ``not to add'' complex designs, with strong motivations and clear experimental supports.
ViT is proposed for image classification.
To adapt ViT to dense vision tasks, recent works choose ``to add'' more CNN-like designs (multi-scale, double channels, spatial reduction). But these add-ons mainly follow the success of CNNs, and their compatibility with attention layers is not verified.
However, our detailed study shows that CNN-like designs are not prerequisites for ViT, and a vanilla ViT architecture plus a better scaling rule (depth, width, input size) and a progressive attention widow strategy can indeed achieve a high detection performance.
Our proposed UViT architectures achieve strong performance on both COCO object detection and instance segmentation. Our uniform design has the potential of supporting multi-modal/multi-task learning and vision-language problems. Most importantly, we hope our work could bring the attention to the community that ViTs may require careful and special architecture design on dense prediction tasks, instead of directly adopting CNN design conventions in black-box.

\clearpage
%
%
\bibliographystyle{splncs04}
\bibliography{egbib}

\appendix

\section{Pascal VOC semantic segmentation}

\paragraph{Settings} Semantic segmentation experiments are conducted on Pascal VOC 2012, which contains 20 foreground classes and 1 background.  For training, we use an augmented version of the dataset~\cite{hariharan2011semantic} with extra annotations of 10582 images (trainaug). The default training setup uses scale jittering of [0.5, 2.0] and random horizontal image flipping.

\subsection{Scaling rule of UViTs} \label{sec:scaling}

\begin{figure*}[h!]
\centering
\vspace{-0.5em}
\includegraphics[scale=0.48]{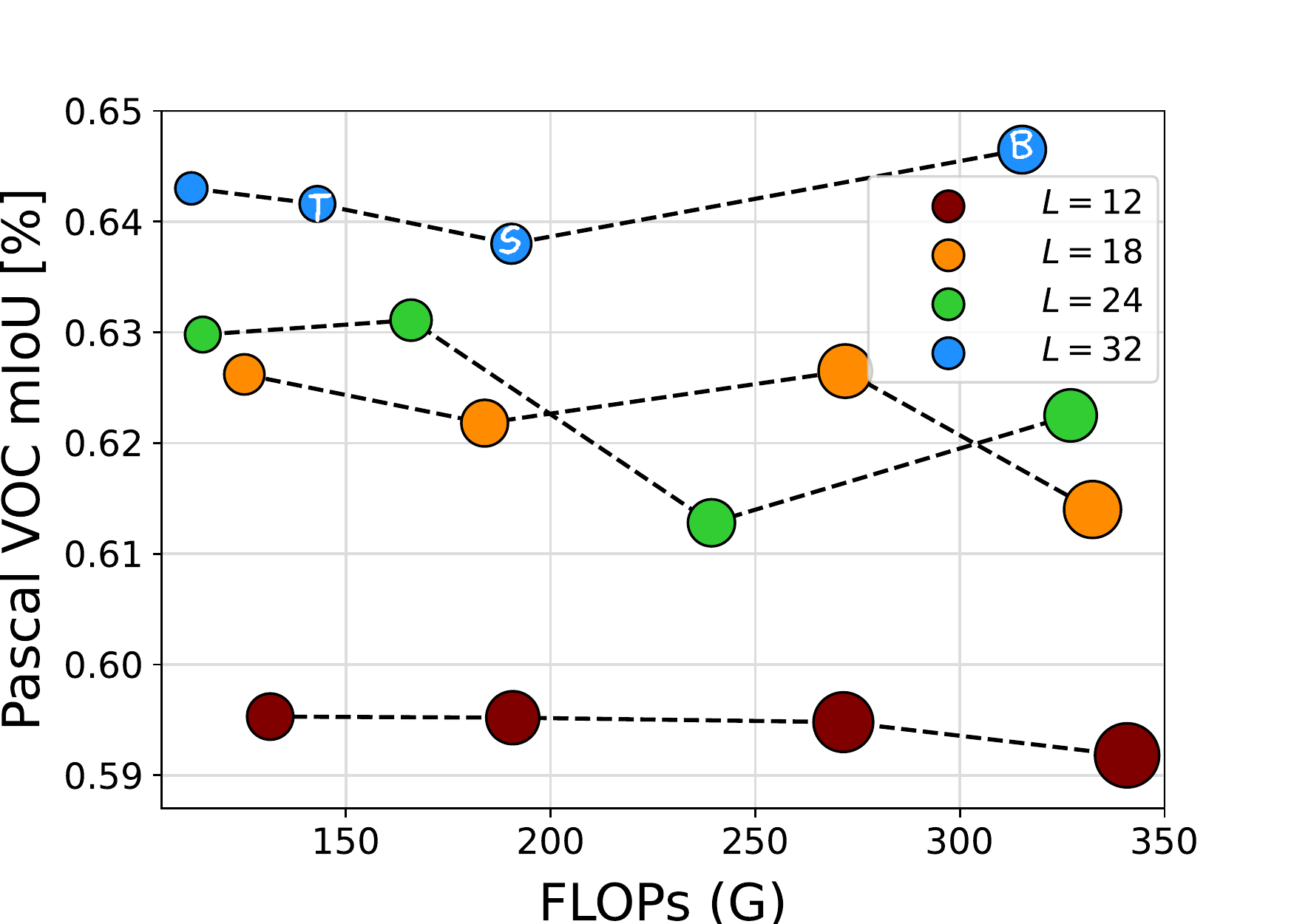}
\includegraphics[scale=0.48]{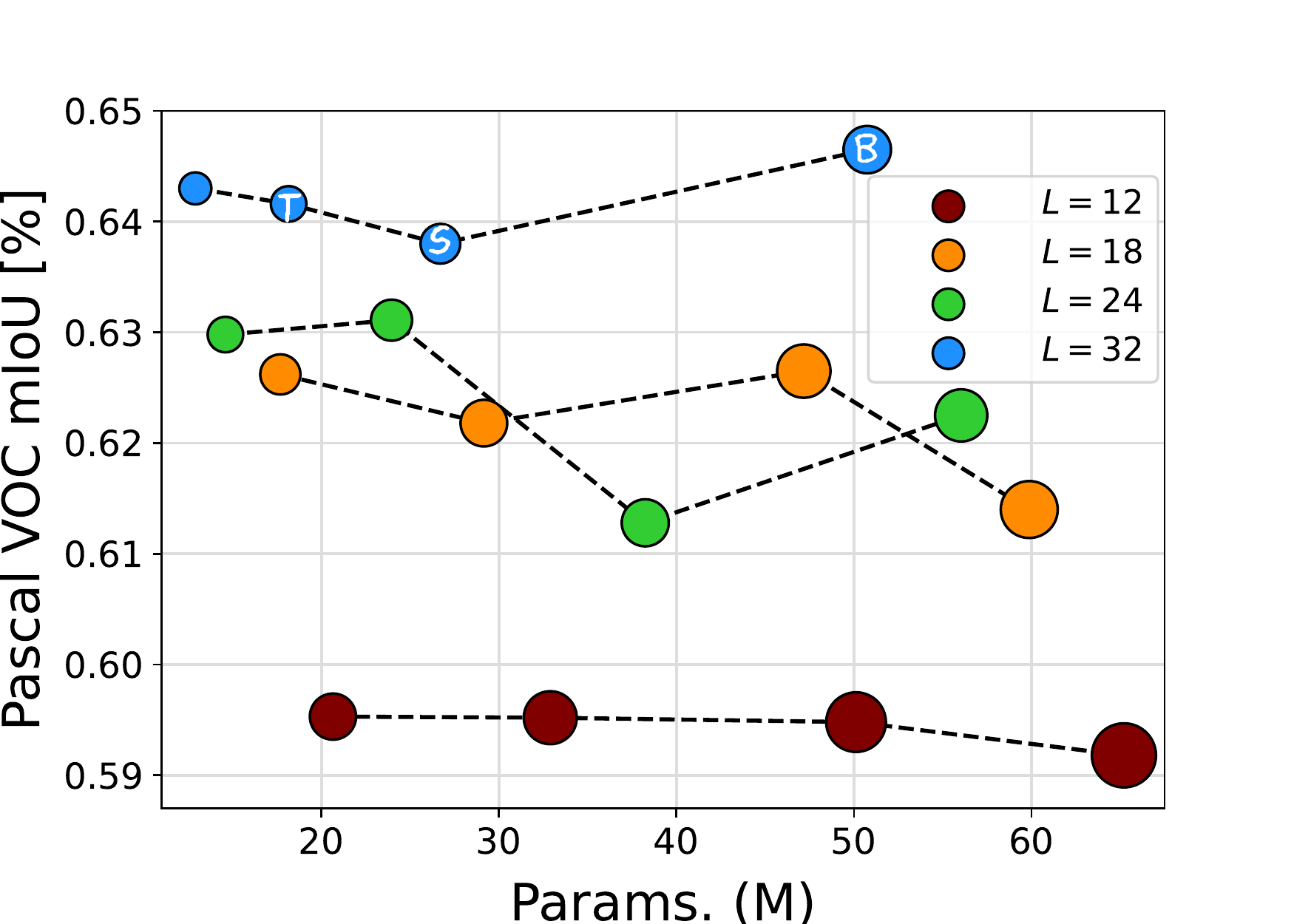}
\centering
\captionsetup{font=small}
\vspace{-0.5em}
\caption{\textbf{Model} scaling rule for UViT on Pascal VOC semantic segmentation (ImageNet pretrained, before COCO pretraining). \textbf{32 attention blocks} (\textcolor{blue}{blue}) perform better than shallower UViTs.
Different sizes of markers represent the hidden sizes (widths).} \label{fig:scaling_rule_model_VOC}
\vspace{-0.5em}
\end{figure*}

To also achieve the best performance-efficiency trade-off on semantic segmentation task, we further systematically study the model scaling of UViTs on depths and widths on the Pascal VOC dataset. We show our results\footnote{This scaling rule is studied before we study the attention window strategy in Section~\ref{sec:attention_window_voc}. Thus for all models in Figure~\ref{fig:scaling_rule_model_VOC} we adopt the window scale as $\frac{1}{2}$, for fair comparisons.} in Figure~\ref{fig:scaling_rule_model_VOC}.
For all models (circle markers), we first train them on ImageNet-1k, then directly fine-tune them on Pascal VOC. We fix the input size as $512\times 512$.
\begin{itemize}[leftmargin=*]
    \item Depth (number of attention blocks): we study different UViT models of depths selected from $\{12, 18, 24, 32\}$.
    \item Width (i.e. hidden size, or output dimension of attention blocks): we will tune the width to further control different model sizes and computation costs to make different scaling rules fairly comparable.
\end{itemize}
In summary, based on our compound scaling rule, we find the UViT of depth 32 performs the best on Pascal VOC.

\subsection{Architectures}

We propose three variants of our UViT variants.
The architecture configurations of our model variants are listed in Table~\ref{table:architecture_VOC}, and are also annotated in Figure~\ref{fig:scaling_rule_model_VOC} (``T'', ``S'', ``B'' in white).
The number of heads is fixed as six, and the expansion ratio of each FFN (feed-forward network) layer is fixed as four in all experiments. 
We also scale up our UViT into a huge version following the design in~\cite{dosovitskiy2020image,zhai2021scaling}, and denote it as ``UViT-H''.

\begin{table}[h!]
\centering
\small
\captionsetup{font=normal}
\setlength{\tabcolsep}{1.8pt}
\caption{Architecture variants of our UViT for Pascal VOC semantic segmentation.}
\resizebox{0.45\textwidth}{!}{
\begin{tabular}{cccc}
\toprule
Name & Depth & \begin{tabular}{@{}c@{}}Hidden\\Size\end{tabular} & Params. (M) \\ \midrule
UViT-T & 32 & 192 & 18.1\\
UViT-S & 32 & 240 & 26.7\\
UViT-B & 32 & 342 & 50.7\\
UViT-H & 32 & 1280 & 529.9\\ \bottomrule
\end{tabular}}\label{table:architecture_VOC}
\end{table}

\subsection{Attention windows on Pascal VOC} \label{sec:attention_window_voc}

In this section, we further study the attention window strategy on the Pascal VOC dataset, using our UViT-B.
As shown in Table~\ref{table:window_ablation_voc},
progressive attention windows again achieve the best performance.
Global attentions in deep layers are vital, and a smaller window in early attentions can improve efficiency.
In conclusion, we set the window scale of our UViT as
``$[2^{-1}]\times 28 \rightarrow [1^{-1}]\times 4$''
for its reduced computation cost.

\begin{table*}[h!]
\small
\caption{Local attention windows in early layers can improve model efficiency, and global attention windows in deep layers are vital to the final performance on Pascal VOC. Model: UViT-B.}
\centering
\resizebox{0.55\textwidth}{!}{
\setlength{\tabcolsep}{1.5pt}
\begin{tabular}{c  c  c}
\toprule
$[\mathrm{window\_scale}] \times \#\mathrm{layers}$ & GFLOPs & \text{AP$_\text{val}$}  \\ \midrule
$[1]\times 32$ & 596.7 & 81.1 \\
$[2^{-1}]\times 32$ & 315.2 & 80.6 \\
$[2^{-1}]\times 28 \rightarrow [1^{-1}]\times 4$ & 350.4 & 81.2 \\
\bottomrule
\end{tabular}
}
\label{table:window_ablation_voc}
\end{table*}

\subsection{Final performance on Pascal VOC}

Following the same procedure in~\cite{chen2014semantic,ronneberger2015u,ghiasi2016laplacian,chen2017deeplab,chen2017rethinking,amirul2017gated,peng2017large,zhao2017pyramid,lin2017refinenet,wang2018understanding}, we employ the
COCO dataset~\cite{lin2014microsoft} to pretrain our model.
From Table~\ref{table:seg} we can see that our UViT is highly compact in terms of the number of parameters, and achieve competitive mIoU with comparable FLOPs.
In addition, it is worth noting that since we only leverage a single-scale feature map, we do not adopt advanced segmentation decoders like ASPP~\cite{chen2017deeplab} but just use plain convolutional layers for predictions, which is to our disadvantage.
Again, without adopting any design conventions from CNNs, the results of our simple UViT architectures suggest that, a simple single-scale transformer backbone can fulfill the dense prediction tasks.

\begin{table}[h]\centering
\captionsetup{font=small}
\caption{Segmentation results on Pascal VOC 2012.
Our UViT leverages a plain convolutional segmentation head, without any test-time augmentation.}
\resizebox{0.9\textwidth}{!}{
\setlength{\tabcolsep}{1.5pt}
\begin{tabular}{cc  cc  cc}
\toprule
   \multicolumn{1}{c}{Backbone} & Resolution & \text{GFLOPs} & \text{Params. (M)}
 & \hspace{2mm}\text{mIoU} \\
    \midrule
    WASPnet-CRF~\cite{artacho2019waterfall} & - & - & 47.5 & 80.4 \\
    DeepLabv3$+$ (ResNet-101)~\cite{chen2018encoder} & 512$\times$512 & 298 & 58.6 & 79.4 \\
    \midrule
    UViT-T (ours) & 512$\times$512 & 163 & 18.1 & 79.0 \\
    UViT-S (ours) & 512$\times$512 & 215 & 26.7 & 79.9 \\
    UViT-B (ours) & 512$\times$512 & 350 & 50.7 & 81.2 \\
    UViT-H (ours) & 640$\times$640 & 3846 & 529.9 & 88.1 \\
    \bottomrule
\end{tabular}}\label{table:seg}
\end{table}

\newpage

\section{Model architectures studied in Figure~\ref{fig:ms_ablation}} \label{sup:arch_details_ablation}

We show details of all architectures studied in Figure~\ref{fig:ms_ablation} (main body) in Table~\ref{table:fig2} below.
As mentioned in Section~\ref{sec:motivation} (main body), we study all combinations of the above three techniques (spatial downsampling ``SD'', multi-scale features ``MF'', doubled channels ``$2\times$''), i.e. eight settings in total, and show the results in Figure~\ref{fig:ms_ablation}. Each dot in Figure~\ref{fig:ms_ablation} indicates an individually designed and trained model. To make all comparisons fair, we carefully design all models such that they are all of around 72 million parameters. We control their FLOPs by changing the depths or attention windows allocated to different stages.

\begin{table*}[b]\centering
\small
\caption{Model architectures in Figure 2 (main body), all studied under a $640\times 640$ input size on MS-COCO. ``SD'': spatial downsampling. ``MF'': multi-scale features. ``$2\times$'': doubled channels. Without any of these three techniques (first section in this table), the whole network has a constant feature resolution and hidden size; all other seven settings below will split the network into three stages, since they require either a progressive feature downsampling or multi-scale features from each stage. Input scale is relative to the 2D shape of the input image $H\times W$ (e.g. $8^{-1}$ indicates the 2D shape of the UViT's sequence feature is $\frac{1}{8}H \times \frac{1}{8}W$). The window scale is relative to the 2D shape of sequence feature's $h\times w$ (e.g. $8^{-1}$ indicates the 2D shape of the attention window is $\frac{1}{8}h \times \frac{1}{8}w$). \ul{Numbers with underscores} in the column ``Output Scale'' indicate feature maps that will be fed into the FPN detection head (i.e., the last output of backbone if no ``MF'' is applied, or features from all three stages if ``MF'' is applied).}
\resizebox{1\textwidth}{!}{
\setlength{\tabcolsep}{3pt}
\begin{tabular}{ccc|ccccccccccccccc|ccc}
\toprule
SD & MF & $2\times$ & \multicolumn{3}{c|}{Input Scale} & \multicolumn{3}{c|}{\#Layers} & \multicolumn{3}{c|}{Window Scale} & \multicolumn{3}{c|}{Hidden Size} & \multicolumn{3}{c|}{Output Scale} & Params. (M) & FLOPs (G) & mAP \\ \midrule
& & & \multicolumn{3}{c}{\multirow{5}{*}{$8^{-1}$}} & \multicolumn{3}{c}{\multirow{5}{*}{18}} & \multicolumn{3}{c}{$16^{-1}$} & \multicolumn{3}{c}{\multirow{5}{*}{384}} & \multicolumn{3}{c}{\multirow{5}{*}{\ul{$8^{-1}$}}} & \multirow{5}{*}{72.1} & 534.1 & 44.5 \\
& & & \multicolumn{3}{c}{} & \multicolumn{3}{c}{} & \multicolumn{3}{c}{$8^{-1}$} & \multicolumn{3}{c}{} & \multicolumn{3}{c}{} &  & 540.9 & 48.2 \\
& & & \multicolumn{3}{c}{} & \multicolumn{3}{c}{} & \multicolumn{3}{c}{$4^{-1}$} & \multicolumn{3}{c}{} & \multicolumn{3}{c}{} &  & 567.9 & 50.1 \\
& & & \multicolumn{3}{c}{} & \multicolumn{3}{c}{} & \multicolumn{3}{c}{$2^{-1}$} & \multicolumn{3}{c}{} & \multicolumn{3}{c}{} &  & 676.2 & 50.7 \\
& & & \multicolumn{3}{c}{} & \multicolumn{3}{c}{} & \multicolumn{3}{c}{$1$} & \multicolumn{3}{c}{} & \multicolumn{3}{c}{} &  & 1109.1 & 50.8 \\
\midrule[1pt]

\multirow{3}{*}{SD} & \multirow{3}{*}{MF} & \multirow{3}{*}{$2\times$} & \multicolumn{5}{|c|}{Stage 1} & \multicolumn{5}{c|}{Stage 2} & \multicolumn{5}{c|}{Stage 3} &
\multirow{3}{*}{Params. (M)} & \multirow{3}{*}{FLOPs (G)} & \multirow{3}{*}{mAP}
\\ \cmidrule{4-8} \cmidrule{9-13} \cmidrule{14-18}
& &  & \begin{tabular}{@{}c@{}}Input\\Scale\end{tabular}& \#Layers & \begin{tabular}{@{}c@{}}Window\\Scale\end{tabular} & \begin{tabular}{@{}c@{}}Hidden\\Size\end{tabular} & \multicolumn{1}{c|}{\begin{tabular}{@{}c@{}}Output\\Scale\end{tabular}} & \begin{tabular}{@{}c@{}}Input\\Scale\end{tabular}& \#Layers & \begin{tabular}{@{}c@{}}Window\\Scale\end{tabular} & \begin{tabular}{@{}c@{}}Hidden\\Size\end{tabular} & \multicolumn{1}{c|}{\begin{tabular}{@{}c@{}}Output\\Scale\end{tabular}} & \begin{tabular}{@{}c@{}}Input\\Scale\end{tabular}& \#Layers & \begin{tabular}{@{}c@{}}Window\\Scale\end{tabular} & \begin{tabular}{@{}c@{}}Hidden\\Size\end{tabular} & \begin{tabular}{@{}c@{}}Output\\Scale\end{tabular} &  &  &  \\ \midrule

$\checkmark$ & &  & \multirow{5}{*}{$8^{-1}$} & 6 & \multirow{5}{*}{1} & \multirow{5}{*}{384} & \multicolumn{1}{c|}{\multirow{5}{*}{$8^{-1}$}} & \multirow{5}{*}{$16^{-1}$} & 6 & \multirow{5}{*}{1} & \multirow{5}{*}{384} & \multicolumn{1}{c|}{\multirow{5}{*}{$16^{-1}$}} & \multirow{5}{*}{$32^{-1}$} & 6 & \multirow{5}{*}{1} & \multirow{5}{*}{384} & \multirow{5}{*}{\ul{$32^{-1}$}} & \multirow{5}{*}{72.1} & 607.1 & 41.0 \\
$\checkmark$ & &  &  & 8 & \multirow{5}{*}{} & \multirow{5}{*}{} & \multicolumn{1}{c|}{\multirow{5}{*}{}} & \multirow{5}{*}{} & 5 & \multirow{5}{*}{} & \multirow{5}{*}{} & \multicolumn{1}{c|}{\multirow{5}{*}{}} & \multirow{5}{*}{} & 5 & \multirow{5}{*}{} & \multirow{5}{*}{} & \multirow{5}{*}{} &  & 688.28 & 42.0 \\
$\checkmark$ & &  &  & 10 & \multirow{5}{*}{} & \multirow{5}{*}{} & \multicolumn{1}{c|}{\multirow{5}{*}{}} & \multirow{5}{*}{} & 4 & \multirow{5}{*}{} & \multirow{5}{*}{} & \multicolumn{1}{c|}{\multirow{5}{*}{}} & \multirow{5}{*}{} & 4 & \multirow{5}{*}{} & \multirow{5}{*}{} & \multirow{5}{*}{} &  & 769.47 & 42.6 \\
$\checkmark$ & &  &  & 12 & \multirow{5}{*}{} & \multirow{5}{*}{} & \multicolumn{1}{c|}{\multirow{5}{*}{}} & \multirow{5}{*}{} & 3 & \multirow{5}{*}{} & \multirow{5}{*}{} & \multicolumn{1}{c|}{\multirow{5}{*}{}} & \multirow{5}{*}{} & 3 & \multirow{5}{*}{} & \multirow{5}{*}{} & \multirow{5}{*}{} &  & 850.68 & 43.0 \\
$\checkmark$ & &  &  & 14 & \multirow{5}{*}{} & \multirow{5}{*}{} & \multicolumn{1}{c|}{\multirow{5}{*}{}} & \multirow{5}{*}{} & 2 & \multirow{5}{*}{} & \multirow{5}{*}{} & \multicolumn{1}{c|}{\multirow{5}{*}{}} & \multirow{5}{*}{} & 2 & \multirow{5}{*}{} & \multirow{5}{*}{} & \multirow{5}{*}{} &  & 931.88 & 43.4 \\ \midrule

 & $\checkmark$ &  & \multirow{5}{*}{$8^{-1}$} & \multirow{5}{*}{6} & $16^{-1}$ & \multirow{5}{*}{384} & \multicolumn{1}{c|}{\multirow{5}{*}{\ul{$8^{-1}$}}} & \multirow{5}{*}{$8^{-1}$} & \multirow{5}{*}{6} & $16^{-1}$ & \multirow{5}{*}{384} & \multicolumn{1}{c|}{\multirow{5}{*}{\ul{$16^{-1}$}}} & \multirow{5}{*}{$8^{-1}$} & \multirow{5}{*}{6} & $16^{-1}$ & \multirow{5}{*}{384} & \multirow{5}{*}{\ul{$32^{-1}$}} & \multirow{5}{*}{72.1} & 534.3 & 44.3 \\
 & $\checkmark$ &  &  & \multirow{5}{*}{} & $8^{-1}$ & \multirow{5}{*}{} & \multicolumn{1}{c|}{\multirow{5}{*}{}} & \multirow{5}{*}{} & \multirow{5}{*}{} & $8^{-1}$ & \multirow{5}{*}{} & \multicolumn{1}{c|}{\multirow{5}{*}{}} & \multirow{5}{*}{} & \multirow{5}{*}{} & $8^{-1}$ & \multirow{5}{*}{} & \multirow{5}{*}{} &  & 541.03 & 47.6 \\
 & $\checkmark$ &  &  & \multirow{5}{*}{} & $4^{-1}$ & \multirow{5}{*}{} & \multicolumn{1}{c|}{\multirow{5}{*}{}} & \multirow{5}{*}{} & \multirow{5}{*}{} & $4^{-1}$ & \multirow{5}{*}{} & \multicolumn{1}{c|}{\multirow{5}{*}{}} & \multirow{5}{*}{} & \multirow{5}{*}{} & $4^{-1}$ & \multirow{5}{*}{} & \multirow{5}{*}{} &  & 568.09 & 49.4 \\
 & $\checkmark$ &  &  & \multirow{5}{*}{} & $2^{-1}$ & \multirow{5}{*}{} & \multicolumn{1}{c|}{\multirow{5}{*}{}} & \multirow{5}{*}{} & \multirow{5}{*}{} & $2^{-1}$ & \multirow{5}{*}{} & \multicolumn{1}{c|}{\multirow{5}{*}{}} & \multirow{5}{*}{} & \multirow{5}{*}{} & $2^{-1}$ & \multirow{5}{*}{} & \multirow{5}{*}{} &  & 676.33 & 50.3 \\
 & $\checkmark$ &  &  & \multirow{5}{*}{} & 1 & \multirow{5}{*}{} & \multicolumn{1}{c|}{\multirow{5}{*}{}} & \multirow{5}{*}{} & \multirow{5}{*}{} & 1 & \multirow{5}{*}{} & \multicolumn{1}{c|}{\multirow{5}{*}{}} & \multirow{5}{*}{} & \multirow{5}{*}{} & 1 & \multirow{5}{*}{} & \multirow{5}{*}{} &  & 1109.3 & 50.2 \\ \midrule

 &  & $\checkmark$ & \multirow{5}{*}{$8^{-1}$} & \multirow{5}{*}{6} & $16^{-1}$ & \multirow{5}{*}{152} & \multicolumn{1}{c|}{\multirow{5}{*}{$8^{-1}$}} & \multirow{5}{*}{$8^{-1}$} & \multirow{5}{*}{6} & $16^{-1}$ & \multirow{5}{*}{304} & \multicolumn{1}{c|}{\multirow{5}{*}{$8^{-1}$}} & \multirow{5}{*}{$8^{-1}$} & \multirow{5}{*}{6} & $16^{-1}$ & \multirow{5}{*}{608} & \multirow{5}{*}{\ul{$8^{-1}$}} & \multirow{5}{*}{73.8} & 558.4 & 43.4 \\
 &  & $\checkmark$ &  & \multirow{5}{*}{} & $8^{-1}$ & \multirow{5}{*}{} & \multicolumn{1}{c|}{\multirow{5}{*}{}} & \multirow{5}{*}{} & \multirow{5}{*}{} & $8^{-1}$ & \multirow{5}{*}{} & \multicolumn{1}{c|}{\multirow{5}{*}{}} & \multirow{5}{*}{} & \multirow{5}{*}{} & $8^{-1}$ & \multirow{5}{*}{} & \multirow{5}{*}{} &  & 561.5 & 44.4 \\
 &  & $\checkmark$ &  & \multirow{5}{*}{} & $4^{-1}$ & \multirow{5}{*}{} & \multicolumn{1}{c|}{\multirow{5}{*}{}} & \multirow{5}{*}{} & \multirow{5}{*}{} & $4^{-1}$ & \multirow{5}{*}{} & \multicolumn{1}{c|}{\multirow{5}{*}{}} & \multirow{5}{*}{} & \multirow{5}{*}{} & $4^{-1}$ & \multirow{5}{*}{} & \multirow{5}{*}{} &  & 587.7 & 46.3 \\
 &  & $\checkmark$ &  & \multirow{5}{*}{} & $2^{-1}$ & \multirow{5}{*}{} & \multicolumn{1}{c|}{\multirow{5}{*}{}} & \multirow{5}{*}{} & \multirow{5}{*}{} & $2^{-1}$ & \multirow{5}{*}{} & \multicolumn{1}{c|}{\multirow{5}{*}{}} & \multirow{5}{*}{} & \multirow{5}{*}{} & $2^{-1}$ & \multirow{5}{*}{} & \multirow{5}{*}{} &  & 692.2 & 46.6 \\
 &  & $\checkmark$ &  & \multirow{5}{*}{} & 1 & \multirow{5}{*}{} & \multicolumn{1}{c|}{\multirow{5}{*}{}} & \multirow{5}{*}{} & \multirow{5}{*}{} & 1 & \multirow{5}{*}{} & \multicolumn{1}{c|}{\multirow{5}{*}{}} & \multirow{5}{*}{} & \multirow{5}{*}{} & 1 & \multirow{5}{*}{} & \multirow{5}{*}{} &  & 1110.2 & 48.3 \\ \midrule

$\checkmark$ & $\checkmark$ &  & \multirow{7}{*}{$8^{-1}$} & 2 & \multirow{7}{*}{1} & \multirow{7}{*}{384} & \multicolumn{1}{c|}{\multirow{7}{*}{\ul{$8^{-1}$}}} & \multirow{7}{*}{$16^{-1}$} & 8 & \multirow{7}{*}{1} & \multirow{7}{*}{384} & \multicolumn{1}{c|}{\multirow{7}{*}{\ul{$16^{-1}$}}} & \multirow{7}{*}{$32^{-1}$} & 8 & \multirow{7}{*}{1} & \multirow{7}{*}{384} & \multirow{7}{*}{\ul{$32^{-1}$}} & \multirow{7}{*}{72.1} & 459.7 & 45.8 \\
$\checkmark$ & $\checkmark$ &  &  & 4 & \multirow{5}{*}{} & \multirow{5}{*}{} & \multicolumn{1}{c|}{\multirow{5}{*}{}} & \multirow{5}{*}{} & 7 & \multirow{5}{*}{} & \multirow{5}{*}{} & \multicolumn{1}{c|}{\multirow{5}{*}{}} & \multirow{5}{*}{} & 7 & \multirow{5}{*}{} & \multirow{5}{*}{} & \multirow{5}{*}{} &  & 540.9 & 47.5 \\
$\checkmark$ & $\checkmark$ &  &  & 6 & \multirow{5}{*}{} & \multirow{5}{*}{} & \multicolumn{1}{c|}{\multirow{5}{*}{}} & \multirow{5}{*}{} & 6 & \multirow{5}{*}{} &  & \multicolumn{1}{c|}{\multirow{5}{*}{}} & \multirow{5}{*}{} & 6 & \multirow{5}{*}{} & \multirow{5}{*}{} & \multirow{5}{*}{} &  & 622.1 & 48.5 \\
$\checkmark$ & $\checkmark$ &  &  & 8 & \multirow{5}{*}{} & \multirow{5}{*}{} & \multicolumn{1}{c|}{\multirow{5}{*}{}} & \multirow{5}{*}{} & 5 & \multirow{5}{*}{} & \multirow{5}{*}{} & \multicolumn{1}{c|}{\multirow{5}{*}{}} & \multirow{5}{*}{} & 5 & \multirow{5}{*}{} & \multirow{5}{*}{} & \multirow{5}{*}{} &  & 703.3 & 48.0 \\
$\checkmark$ & $\checkmark$ &  &  & 10 & \multirow{5}{*}{} & \multirow{5}{*}{} & \multicolumn{1}{c|}{\multirow{5}{*}{}} & \multirow{5}{*}{} & 4 & \multirow{5}{*}{} & \multirow{5}{*}{} & \multicolumn{1}{c|}{\multirow{5}{*}{}} & \multirow{5}{*}{} & 4 & \multirow{5}{*}{} & \multirow{5}{*}{} & \multirow{5}{*}{} &  & 784.5 & 48.6 \\
$\checkmark$ & $\checkmark$ &  &  & 12 & \multirow{5}{*}{} & \multirow{5}{*}{} & \multicolumn{1}{c|}{\multirow{5}{*}{}} & \multirow{5}{*}{} & 3 & \multirow{5}{*}{} & \multirow{5}{*}{} & \multicolumn{1}{c|}{\multirow{5}{*}{}} & \multirow{5}{*}{} & 3 & \multirow{5}{*}{} & \multirow{5}{*}{} & \multirow{5}{*}{} &  & 865.7 & 50.2 \\
$\checkmark$ & $\checkmark$ &  &  & 15 & \multirow{5}{*}{} & \multirow{5}{*}{} & \multicolumn{1}{c|}{\multirow{5}{*}{}} & \multirow{5}{*}{} & 2 & \multirow{5}{*}{} & \multirow{5}{*}{} & \multicolumn{1}{c|}{\multirow{5}{*}{}} & \multirow{5}{*}{} & 1 & \multirow{5}{*}{} & \multirow{5}{*}{} & \multirow{5}{*}{} &  & 989.5 & 50.4 \\
\midrule

$\checkmark$ & & $\checkmark$ & \multirow{5}{*}{$8^{-1}$} & \multirow{5}{*}{16} & \multirow{5}{*}{1} & 128 & \multicolumn{1}{c|}{\multirow{5}{*}{$8^{-1}$}} & \multirow{5}{*}{$16^{-1}$} & \multirow{5}{*}{1} & \multirow{5}{*}{1} & 256 & \multicolumn{1}{c|}{\multirow{5}{*}{$16^{-1}$}} & \multirow{5}{*}{$32^{-1}$} & 9 & \multirow{5}{*}{1} & 512 & \multirow{5}{*}{\ul{$32^{-1}$}} & 70.2 & 529.1 & 37.6 \\
$\checkmark$ & & $\checkmark$ &  & \multirow{5}{*}{} & \multirow{5}{*}{} & 160 & \multicolumn{1}{c|}{\multirow{5}{*}{}} & \multirow{5}{*}{} & \multirow{5}{*}{} & \multirow{5}{*}{} & 320 & \multicolumn{1}{c|}{\multirow{5}{*}{}} & \multirow{5}{*}{} & 5 & \multirow{5}{*}{} & 640 & \multirow{5}{*}{} & 69.3 & 581.7 & 38.9 \\
$\checkmark$ & & $\checkmark$ &  & \multirow{5}{*}{} & \multirow{5}{*}{} & 192 & \multicolumn{1}{c|}{\multirow{5}{*}{}} & \multirow{5}{*}{} & \multirow{5}{*}{} & \multirow{5}{*}{} & 384 & \multicolumn{1}{c|}{\multirow{5}{*}{}} & \multirow{5}{*}{} & 3 & \multirow{5}{*}{} & 768 & \multirow{5}{*}{} & 69.3 & 637.4 & 40.2 \\
$\checkmark$ & & $\checkmark$ &  & \multirow{5}{*}{} & \multirow{5}{*}{} & 224 & \multicolumn{1}{c|}{\multirow{5}{*}{}} & \multirow{5}{*}{} & \multirow{5}{*}{} & \multirow{5}{*}{} & 448 & \multicolumn{1}{c|}{\multirow{5}{*}{}} & \multirow{5}{*}{} & 2 & \multirow{5}{*}{} & 896 & \multirow{5}{*}{} & 71.4 & 696.6 & 41.7 \\
$\checkmark$ & & $\checkmark$ &  & \multirow{5}{*}{} & \multirow{5}{*}{} & 256 & \multicolumn{1}{c|}{\multirow{5}{*}{}} & \multirow{5}{*}{} & \multirow{5}{*}{} & \multirow{5}{*}{} & 512 & \multicolumn{1}{c|}{\multirow{5}{*}{}} & \multirow{5}{*}{} & 1 & \multirow{5}{*}{} & 1024 & \multirow{5}{*}{} & 69.2 & 756.5 & 42.5 \\ \midrule

 & $\checkmark$ & $\checkmark$ & \multirow{4}{*}{$8^{-1}$} & \multirow{4}{*}{6} & $16^{-1}$ & \multirow{4}{*}{152} & \multicolumn{1}{c|}{\multirow{4}{*}{\ul{$8^{-1}$}}} & \multirow{4}{*}{$8^{-1}$} & \multirow{4}{*}{6} & $16^{-1}$ & \multirow{4}{*}{304} & \multicolumn{1}{c|}{\multirow{4}{*}{\ul{$16^{-1}$}}} & \multirow{4}{*}{$8^{-1}$} & \multirow{4}{*}{6} & $16^{-1}$ & \multirow{4}{*}{608} & \multirow{4}{*}{\ul{$32^{-1}$}} & \multirow{4}{*}{73.8} & 566.3 & 45.7 \\
 & $\checkmark$ & $\checkmark$ &  & \multirow{4}{*}{} & $8^{-1}$ & \multirow{4}{*}{} & \multicolumn{1}{c|}{\multirow{5}{*}{}} & \multirow{4}{*}{} & \multirow{4}{*}{} & $8^{-1}$ & \multirow{4}{*}{} & \multicolumn{1}{c|}{\multirow{5}{*}{}} & \multirow{4}{*}{} & \multirow{4}{*}{} & $8^{-1}$ & \multirow{4}{*}{} & \multirow{4}{*}{} &  & 569.5 & 46.4 \\
 & $\checkmark$ & $\checkmark$ &  & \multirow{4}{*}{} & $4^{-1}$ & \multirow{4}{*}{} & \multicolumn{1}{c|}{\multirow{5}{*}{}} & \multirow{4}{*}{} & \multirow{4}{*}{} & $4^{-1}$ & \multirow{4}{*}{} & \multicolumn{1}{c|}{\multirow{5}{*}{}} & \multirow{4}{*}{} & \multirow{4}{*}{} & $4^{-1}$ & \multirow{4}{*}{} & \multirow{4}{*}{} &  & 595.6 & 48.1 \\
 & $\checkmark$ & $\checkmark$ &  & \multirow{4}{*}{} & $2^{-1}$ & \multirow{4}{*}{} & \multicolumn{1}{c|}{\multirow{5}{*}{}} & \multirow{4}{*}{} & \multirow{4}{*}{} & $2^{-1}$ & \multirow{4}{*}{} & \multicolumn{1}{c|}{\multirow{5}{*}{}} & \multirow{4}{*}{} & \multirow{4}{*}{} & $2^{-1}$ & \multirow{4}{*}{} & \multirow{4}{*}{} &  & 700.1 & 49.0 \\ \midrule

$\checkmark$ & $\checkmark$ & $\checkmark$ & \multirow{6}{*}{$8^{-1}$} & 16 & \multirow{6}{*}{1} & 128 & \multicolumn{1}{c|}{\multirow{6}{*}{\ul{$8^{-1}$}}} & \multirow{6}{*}{$16^{-1}$} & \multirow{6}{*}{1} & \multirow{6}{*}{1} & 256 & \multicolumn{1}{c|}{\multirow{6}{*}{\ul{$16^{-1}$}}} & \multirow{6}{*}{$32^{-1}$} & 9 & \multirow{6}{*}{1} & 512 & \multirow{6}{*}{\ul{$32^{-1}$}} & 73.3 & 552.1 & 44.3 \\
$\checkmark$ & $\checkmark$ & $\checkmark$ &  & 16 & \multirow{4}{*}{} & 160 & \multicolumn{1}{c|}{\multirow{5}{*}{}} & \multirow{4}{*}{} & \multirow{4}{*}{} & \multirow{4}{*}{} & 320 & \multicolumn{1}{c|}{\multirow{5}{*}{}} & \multirow{4}{*}{} & 5 & \multirow{4}{*}{} & 640 & \multirow{4}{*}{} & 72.4 & 604.9 & 45.5 \\
$\checkmark$ & $\checkmark$ & $\checkmark$ &  & 16 & \multirow{4}{*}{} & 192 & \multicolumn{1}{c|}{\multirow{5}{*}{}} & \multirow{4}{*}{} & \multirow{4}{*}{} & \multirow{4}{*}{} & 384 & \multicolumn{1}{c|}{\multirow{5}{*}{}} & \multirow{4}{*}{} & 3 & \multirow{4}{*}{} & 768 & \multirow{4}{*}{} & 72.4 & 660.7 & 47.6 \\
$\checkmark$ & $\checkmark$ & $\checkmark$ &  & 16 & \multirow{4}{*}{} & 224 & \multicolumn{1}{c|}{\multirow{5}{*}{}} & \multirow{4}{*}{} & \multirow{4}{*}{} & \multirow{4}{*}{} & 448 & \multicolumn{1}{c|}{\multirow{5}{*}{}} & \multirow{4}{*}{} & 2 & \multirow{4}{*}{} & 896 & \multirow{4}{*}{} & 74.5 & 719.9 & 48.8 \\
$\checkmark$ & $\checkmark$ & $\checkmark$ &  & 16 & \multirow{4}{*}{} & 256 & \multicolumn{1}{c|}{\multirow{5}{*}{}} & \multirow{4}{*}{} & \multirow{4}{*}{} & \multirow{4}{*}{} & 512 & \multicolumn{1}{c|}{\multirow{5}{*}{}} & \multirow{4}{*}{} & 1 & \multirow{4}{*}{} & 1024 & \multirow{4}{*}{} & 72.4 & 779.9 & 49.4 \\
$\checkmark$ & $\checkmark$ & $\checkmark$ &  & 28 & \multirow{4}{*}{} & 224 & \multicolumn{1}{c|}{\multirow{5}{*}{}} & \multirow{4}{*}{} & \multirow{4}{*}{} & \multirow{4}{*}{} & 448 & \multicolumn{1}{c|}{\multirow{5}{*}{}} & \multirow{4}{*}{} & 1 & \multirow{4}{*}{} & 896 & \multirow{4}{*}{} & 72.1 & 992.3 & 49.5 \\

\bottomrule
\end{tabular}
}
\label{table:fig2}
\end{table*}

\section{Model architectures in Figure~\ref{fig:scaling_rule_input_size} and Figure~\ref{fig:scaling_rule_model}} \label{sup:arch_details_scaling}

We show all architectures studied in our compound scaling rule in Figure~\ref{fig:scaling_rule_input_size} and Figure~\ref{fig:scaling_rule_model} (main body). All models are of $2^{-1}$-scale attention windows for fair comparisons.

\begin{table*}[b]\centering
\small
\captionsetup{font=small}
\caption{Model architectures in Figure 4 and Figure 5 (MS-COCO) (in main body). Configurations (depth, width) of UViT-T/S/B are annotated.}
\resizebox{0.72\textwidth}{!}{
\setlength{\tabcolsep}{3pt}
\begin{tabular}{cccccc}
\toprule
Input Size & Depth & Width & Params. (M) & FLOPs (G) & mAP \\ \midrule
\multirow{5}{*}{$640\times 640$} & \multirow{5}{*}{18} & 384 & 72.1 & 676.2 & 50.4 \\
 &  & 432 & 80.9 & 748.3 & 50.5 \\
 &  & 462 & 86.9 & 796.6 & 50.7 \\
 &  & 492 & 93.3 & 847.4 & 50.4 \\
 &  & 564 & 110.2 & 979.4 & 50.1 \\ \midrule
\multirow{6}{*}{$768\times 768$} & \multirow{6}{*}{18} & 288 & 58.2 & 725.9 & 51.1 \\
 &  & 306 & 60.7 & 761.1 & 51.5 \\
 &  & 330 & 64.3 & 810.0 & 51.3 \\
 &  & 384 & 73.1 & 928.5 & 51.5 \\
 &  & 432 & 82.1 & 1043.5 & 51.6 \\
 &  & 462 & 88.2 & 1120.1 & 51.3 \\ \midrule
\multirow{6}{*}{$896\times 896$} & \multirow{6}{*}{18} & 186 & 47.4 & 710.2 & 51 \\
 &  & 222 (UViT-T) & 51.0 & 801.4 & 51.3 \\
 &  & 246 & 53.8 & 866.1 & 51.7 \\
 &  & 288 (UViT-S) & 59.2 & 986.8 & 51.7 \\
 &  & 330 & 65.4 & 1117.1 & 52.1 \\
 &  & 384 (UViT-B) & 74.4 & 1298.7 & 52.3 \\ \midrule
\multirow{7}{*}{$1024\times 1024$} & \multirow{7}{*}{18} & 120 & 42.6 & 710.3 & 47.9 \\
 &  & 132 & 43.5 & 750.1 & 48.9 \\
 &  & 144 & 44.4 & 791.0 & 49.3 \\
 &  & 162 & 45.8 & 854.3 & 50.4 \\
 &  & 198 & 49.3 & 987.6 & 51.4 \\
 &  & 246 & 54.7 & 1179.7 & 51.7 \\
 &  & 288 & 60.3 & 1361.2 & 52.0 \\ \midrule
\multirow{5}{*}{$896\times 896$} & \multirow{5}{*}{12} & 276 & 52.1 & 748.4 & 50.8 \\
 &  & 300 & 54.4 & 796.2 & 50.8 \\
 &  & 324 & 56.9 & 846.2 & 51.0 \\
 &  & 360 & 60.9 & 925.0 & 51.5 \\
 &  & 390 & 64.5 & 994.2 & 51.5 \\ \midrule
\multirow{5}{*}{$896\times 896$} & \multirow{5}{*}{24} & 156 & 46.5 & 739.0 & 50.6 \\
 &  & 180 & 49.2 & 813.8 & 50.8 \\
 &  & 192 & 50.6 & 852.7 & 51.3 \\
 &  & 258 & 60.1 & 1085.4 & 51.8 \\
 &  & 294 & 66.3 & 1225.7 & 51.6 \\ \midrule
\multirow{5}{*}{$896\times 896$} & \multirow{5}{*}{32} & 120 & 44.6 & 732.5 & 50 \\
 &  & 132 & 45.9 & 777.4 & 50.4 \\
 &  & 144 & 47.3 & 823.8 & 51.2 \\
 &  & 180 & 52.3 & 971.1 & 51.5 \\
 &  & 240 & 62.8 & 1244.4 & 52.0 \\ \midrule
\multirow{6}{*}{$896\times 896$} & \multirow{6}{*}{40} & 96 & 43.2 & 723.2 & 48.5 \\
 &  & 102 & 43.8 & 749.3 & 49.1 \\
 &  & 114 & 45.2 & 802.9 & 50.1 \\
 &  & 126 & 46.8 & 858.2 & 50.7 \\
 &  & 150 & 50.3 & 974.0 & 51.2 \\
 &  & 156 & 51.2 & 1004.0 & 51.2 \\
\bottomrule
\end{tabular}
}
\label{table:fig4}
\end{table*}

\end{document}